\documentclass[runningheads]{llncs}
\usepackage{graphicx}
\usepackage{svg}
\usepackage{colortbl}
\usepackage[normalem]{ulem}
\usepackage{comment}
\usepackage{amssymb}
\usepackage{amsmath}
\usepackage{comment}
\usepackage{graphicx}
\usepackage{appendix}
\usepackage{tabularx}
\usepackage[normalem]{ulem}
\useunder{\uline}{\ul}{}
\usepackage{float}

\usepackage{booktabs}
\begin{document}
\bibliographystyle{unsrt}

\title{CIAMS: Clustering Indices based Automatic classification Model Selection\thanks{Supported by Claritrics Inc. d.b.a BUDDI AI.}}
%
%
\author{Sudarsun Santhiappan\thanks{Corresponding author}\inst{1} \and 
Nitin Shravan\inst{3} \and 
Balaraman Ravindran\inst{1,2} } 
\authorrunning{S. Santhiappan et al.}
%
\institute{Dept. of Computer Science and Engineering, \\ 
Indian Institute of Technology, Madras \\
Chennai, India \\
\and
Robert Bosch Center for Data Science and AI \\
Indian Institute of Technology, Madras \\
Chennai, India \\
\email{\{sudarsun, ravi\}@cse.iitm.ac.in} \and
Claritrics Inc. d.b.a BUDDI AI, New York, NY, USA \\
}

%
\maketitle              

\begin{abstract}
Classification model selection is a process of identifying a suitable model class for a given classification task on a dataset. Traditionally, model selection is based on cross-validation, meta-learning, and user preferences, which are often time-consuming and resource-intensive. The performance of any machine learning classification task depends on the choice of the model class, the learning algorithm, and the dataset's characteristics. Our work proposes a novel method for automatic classification model selection from a set of candidate model classes by determining the empirical model-fitness for a dataset based only on its clustering indices. Clustering Indices measure the ability of a clustering algorithm to induce good quality neighborhoods with similar data characteristics. We propose a regression task for a given model class, where the clustering indices of a given dataset form the features and the dependent variable represents the expected classification performance. We compute the dataset clustering indices and directly predict the expected classification performance using the learned regressor for each candidate model class to recommend a suitable model class for dataset classification. We evaluate our model selection method through cross-validation with 60 publicly available binary class datasets and show that our \emph{top3} model recommendation is accurate for over 45 of 60 datasets. We also propose an end-to-end Automated ML system for data classification based on our model selection method. We evaluate our end-to-end system against popular commercial and noncommercial Automated ML systems using a different collection of 25 public domain binary class datasets. We show that the proposed system outperforms other methods with an excellent average rank of 1.68.
\end{abstract}

\section{Introduction}

Selecting a suitable machine learning model that maximizes the performance measured for a given task is an essential step in data science. The traditional approach is to train different models, evaluate their performance on a validation set, and choose the best model. However, this method is time-consuming and resource-intensive. \emph{Automated machine learning} is an active area of research to automatically select a suitable machine learning model for a given task. Researchers have tried to address the model selection problem through various approaches such as meta-learning \cite{bsc-rlaimat-03:ibl,Vainshtein2018AHA:autodi,10.1145/3357384.3357896:autogrd}, deep reinforcement learning \cite{DBLP:journals/corr/abs-1905-10345:alphad3m}, Bayesian optimization \cite{thornton2013auto:auto-weka,NIPS2015_5872:auto-sklearn}, evolutionary algorithms \cite{inbook:tpot,DBLP:journals/corr/abs-1803-00684:autostacker,Real2020AutoMLZeroEM:automl-zero}, and budget-based evaluation~\cite{Li2017EfficientHO:budget}.    

Analyzing the data characteristics is essential for selecting an appropriate classification model and feature engineering. However, supposing we can estimate the empirical classification model performance with explainability for the dataset apriori, it becomes straightforward to pick a suitable classifier model class to solve the problem. This setup is advantageous while working with large datasets, as it is laborious and time-consuming to evaluate different classifier model classes for model selection.

Clustering methods group the data points having similar characteristics into neighborhoods or disjuncts of different sizes. Clustering indices \cite{clustercrit} are cluster evaluation metrics used to assess the quality of the clusters induced by a clustering algorithm.  Clustering indices measure the ability of a clustering algorithm to induce good quality neighborhoods with similar data characteristics. We hypothesize that the clustering indices provide a low-dimensional vector representation of the dataset characteristics with respect to a clustering method. When we use different clustering methods to compute the clustering indices, we can generate different views of the dataset characteristics.  Combining multiple views of the data characteristics in terms of clustering indices gives us a rich feature space representation of the dataset characteristics.

We use the term \emph{model-fitness} to denote the ability of a model class to learn a classification task on a given dataset.  The empirical model-fitness of a dataset can be measured based on the expected classification performance of a model class on a dataset. We use the $F_1$ score as the classification performance metric in our experiments, but the idea is agnostic to any metric. The empirical performance of a classifier depends on the ability of the classifier hypothesis to model the data characteristics \cite{Ho2002}. We hypothesize that the dataset characteristics represented by the clustering indices correlate strongly with the empirical model fitness.  



In this paper, we propose \emph{CIAMS}, a novel Clustering Indices based Automatic Model Selection method from a set of model classes by estimating the empirical model fitness for a given binary class dataset from only its clustering indices representation of the data characteristics.  We model the relationship between the clustering indices of a dataset and the empirical model fitness of a model class as a regression problem. We learn a regressor for each model class from a set of model classes on several datasets by randomly drawing subsamples with replacement. Constructing multiple subsamples allows us to increase the number of data points to train our regression model.  Another advantage of using subsamples is to provide broader coverage of the dataset variance characteristic for regression modeling.

We train independent regressors for each model class with the best achievable classification performance for each dataset as the output and its respective estimated clustering indices as the input predictors.  We tune every candidate classifier model for maximum classification performance concerning the dataset subsample.  This way, when the regressors learn the mapping between clustering indices and the maximum achievable classification performance for every model class. The automation model selection process, implemented as a prediction task, can estimate the model fitness directly in terms of the expected classification performance. Using the estimated model fitness, we rank the candidate model classes to suggest the top model classes for a given dataset as the recommendation.  We limit ourselves from suggesting the model class hyper-parameters.  We believe mapping the hyper-parameters to the dataset characteristics is a separate problem, which we mark as one of the future extensions of \emph{CIAMS}.  We validate our model selection regressor through cross-validation using \emph{60 (sixty)} public domain binary class datasets and observe that our model recommendation is accurate for over three-fourths of the datasets. 

We extend our automatic model classification method to an end-to-end \emph{Automated Machine Learning platform} to offer classification modeling as a service. We use the \emph{top3} model classes predicted by our model selection method to build tuned classifiers using the labeled portion of the given \emph{production} dataset.  We define \emph{production} dataset as the input provided by an end-user containing labeled and unlabeled portions, from which we learn these classifiers, followed by predicting the labels for the unlabeled data points. The best-performing model, chosen through cross-validation, among the \emph{top3} tuned classifier models is offered as a service to predict labels for the unlabeled portion of the production dataset.  We validate our platform against other commercial and noncommercial automated machine learning systems using a different set of \emph{25 (twenty-five)} public domain binary class datasets of varied sizes. The comparison experiment shows that our platform outperforms other systems with an excellent average rank of \emph{1.68}, proving its viability in building practical applications.

\vspace{5pt}
The main contributions of this paper are:
\begin{itemize}
    \item A novel hypothesis is that the classification performance of a model class for a dataset is a function of the dataset's clustering indices.
    \item A novel method to predict the expected classification performance of a model class for a dataset without building the classification model. 
    \item A novel application of clustering indices for automatic model selection from a list of model classes for a given dataset.
    \item A novel automated machine learning platform (Automated ML) for learning and deploying a classifier model as a \emph{service}. 

\end{itemize}

We organize the remainder of the paper as follows. Section \ref{background} lists the related techniques and approaches for model selection and model fitness assessment. Section \ref{Summary} summarizes our approach to automatic model selection. Section~\ref{MSsystem} gives a detailed explanation of our proposed model selection system. Section~\ref{expsetup} describes the entire experimental setup and parameter configuration. Section~\ref{results} validates our system and narrates the results obtained from the experimental study. Section \ref{conclusion} provides the concluding remarks and next steps.

\section{Related work}\label{background}

In this section, we summarize various approaches from the literature for automatic model selection organized into different categories. 

\begin{itemize}
    \item \textit{Random Search:} Early research works hypothesized automated model selection as a Combined Algorithm Selection and Hyperparameter optimization (CASH) problem. Amazon's Sagemaker \cite{sagemaker_whitepaper,doi:10.1002/9781119556749.ch16} is an example of a commercial Automated ML platform that follows the CASH paradigm. H$_2$O AutoML~\cite{h2oICML,H2OAutoML}, an open-source Automated ML platform, uses fast random search and ensemble methods like stacking to achieve competitive results. 
    
    \item \textit{Bayesian Optimization:} Auto-weka \cite{thornton2013auto:auto-weka} and Auto-sklearn \cite{NIPS2015_5872:auto-sklearn} are Automated ML frameworks extensions of the popular Weka and Scikit-learn libraries, respectively. Auto-Weka \cite{thornton2013auto:auto-weka} uses a state-of-the-art Bayesian optimization method, random-forest-based Sequential Model-based Algorithm Configuration (SMAC), for automated model selection. Auto-sklearn \cite{NIPS2015_5872:auto-sklearn} builds on top of the Bayesian optimization solution in Auto-weka by including meta-learning for initialization of the Bayesian optimizer and ensembling to provide high predictive performance. Microsoft Azure Automated ML \cite{azureautomatedml,fusi2018probabilistic}
     uses Bayesian optimization and collaborative filtering for automatic model selection and tuning.
    
    \item \textit{Evolutionary Algorithms:} TPOT \cite{inbook:tpot} is a Python-based framework that uses the Genetic Programming algorithm to evolve and optimize tree-based machine learning pipelines. Autostacker \cite{DBLP:journals/corr/abs-1803-00684:autostacker} is similar to TPOT, but stacked layers represent the machine learning pipeline. AutoML-Zero \cite{Real2020AutoMLZeroEM:automl-zero} uses basic mathematical operations as building blocks to discover complete machine learning algorithms through evolutionary algorithms. FLAML \cite{wang2021flaml:flaml} uses an Estimated Cost for Improvement (ECI)-based prioritization to find the optimal learning algorithm in low-cost environments.
    
    \item \textit{Deep Reinforcement Learning:} AlphaD3M \cite{DBLP:journals/corr/abs-1905-10345:alphad3m} uses deep reinforcement learning to synthesize various components in the machine learning pipeline to obtain maximum performance measures.
    
    \item \textit{Meta-Learning:} Brazdil et al. \cite{bsc-rlaimat-03:ibl,Brazdil00rankingclassification} uses a k-nearest neighbor approach based on the dataset characteristics to provide a ranked list of classifiers using different ranking methods based on accuracy and time information. AutoDi~\cite{Vainshtein2018AHA:autodi} uses word-embedding features and dataset meta-features for automatic model selection. AutoGRD \cite{10.1145/3357384.3357896:autogrd} represents the datasets as graphs to extract features for training the meta-learner. AutoClust \cite{autoclust2020} uses clustering indices as meta-features to select suitable clustering algorithms and hyper-parameters automatically.  Sahni et al. \cite{10.1145/3430984.3431029:pappu} developed an approach using meta-features to select a sampling method for imbalanced data classification automatically. Santhiappan et al. \cite{10.1145/3430984.3430997:sudar-cods} propose a method using clustering indices as meta-features to estimate the empirical binary classification complexity of the dataset.
\end{itemize}

Our method follows the meta-learning paradigm, wherein we learn the relationship between the extracted meta-features of the dataset in terms of clustering indices and the expected classification performance of a model class. The trained meta-learner predicts the classification performance of a model class for an unseen dataset without building a classifier model. The differentiation among various meta-learning methods pivots on the choice of the meta-features extracted from the dataset. The following list presents the meta-features from the literature, organized into different categories.

\begin{itemize}
    \item \textit{Statistical \& Information-Theoretic }\cite{Katz2016ExploreKitAF:explorekit,DatametricforDM}: These measures include the number of data points in the dataset, number of classes, number of variables with a numeric and symbolic data type, average and variance of every feature, the entropy of individual features, and more. These metrics capture important meta-information about the dataset. 
    \begin{itemize}
        \item \texttt{Class Boundary:} The nature of the class margin of a dataset is an essential characteristic reflecting its classifiability. Measures such as inter-class~\& intra-class nearest-neighbor distance, error rate, and non-linearity of the nearest-neighbor classifier try to capture the underlying class margin properties such as shape and narrowness between classes.
        \item \texttt{Class Imbalance:} Machine learning methods in their default settings are biased toward learning the majority class due to a lack of data points representing the minority class. Features such as entropy of class proportions and class-imbalance ratio strongly reflect dataset characteristics such as the classification complexity of a dataset. 
        \item \texttt{Data Sparsity }\cite{liling}: Sparse regions in the dataset affect the classifier's learning ability leading to poor performance. The average number of features per dimension, the average number of PCA dimensions per point, and the ratio of PCA dimension to the original dimension capture sparsity in the dataset.
        
    \end{itemize}
    
    \item \textit{Feature-based }\cite{Orriols2010,featurebased1}: The learning ability of methods is highly correlated with the features' discriminatory power. Fisher's discriminant ratio, Overlap region volume, and feature efficiency are among many measures from the literature that tries to capture the ability to learn.  
    
    \item \textit{Model based }\cite{10.5555/647859.736120:hyperparameters,Oddbitesbannanas,higherorderapproach}: The hyper-parameters of a model directly affect the model performance. For instance, in a tree-based model, hyper-parameters such as the number of leaf nodes, maximum depth, and average \textit{gain-ratio} difference serve as meta-features.  Likewise, the number of support vectors required in SVM modeling is a meta-feature. 
   
    \begin{itemize}
        \item \texttt{Linearity }\cite{Orriols2010,Hoekstra1996OnTN}: Most classifiers have high performance when the dataset is linear. To capture the inherent linearity present in data, we use model-based measures, such as the error rate of a linear SVM, and non-linearity of the linear classifier, as meta-features.
        \item \texttt{Landmarking}: Bensusan et al. \cite{article:landmarker1,landmarker2} notes that providing the performance measures obtained using simple learning algorithms (baselines) as a meta-feature has a strong co-relation with the classification performance of the considered algorithm. F{\"u}rnkranz et al. \cite{fuernkranz} explores different landmarking variants like relative landmarking, sub-sample landmarking techniques, and their effectiveness in several learning tasks, such as decision tree pruning.
        \begin{itemize}
            \item Landmarking is one of the most effective methods for meta-learning based automatic model selection \cite{fuernkranz}. Landmarking uses simple classifiers' performance on a dataset to capture the underlying characteristics. Landmarking requires building several simple classifier models on the dataset to extract features. In comparison, our proposed model selection approach requires building dataset clusters to extract clustering indices features.  We compare the performance of landmarking and clustering indices features through an extrinsic regression task in Section \ref{section:comparing-equivalent-methods}.
            \item The computational cost of extracting the clustering indices as the meta-features for big datasets is mitigated through subsampling similar to Landmarkers \cite{fuernkranz}. Petrak et al. \cite{Petrak00fastsubsampling} establish the ``\textit{Similarity of regions of Expertise}" property, which says that the meta-features from several subsamples of the dataset collectively represent the characteristics of the full dataset. We also empirically validate the clustering indices upholding the said property in Section \ref{Whysubsamples}. 
        \end{itemize}
    \end{itemize}

    \item \textit{Graph-based }\cite{Garcia,MORAIS}: Graph representation of a dataset can help extract useful meta-features such as mean network density, coefficient of clustering, and hub score for several meta-learning tasks.

\end{itemize}

Data characterization is a crucial setup in understanding the nuances in a dataset. When specific dataset properties are known, it helps choose the suitable method or algorithm to solve the task. Several meta-features discussed in the literature are targeted at specific dataset characteristics. Enlisting the meta-features of a dataset is a laborious process that is costly in terms of time and computing power. We choose clustering indices to be the meta-features to represent dataset characteristics. Clustering indices are evaluation metrics to estimate how well a clustering algorithm grouped the data with similar characteristics.
Clustering indices are scalar values that indicate the nuances in a dataset under different clustering assumptions. Computing the clustering indices is a parallelizable process whose time complexity is proportional to the size of a dataset subsample. When we generate clustering indices under different clustering assumptions, we get more comprehensive coverage of the data characteristics. 

The clustering indices approach to represent the data characteristics is similar to the landmarking approach, in principle. Despite requiring more computing power for dataset clustering, our experiments in Section~\ref{results} empirically show that the clustering indices capture a richer dataset characteristic representation for providing better generalization. 


\section{Our approach to Automatic Model Selection}\label{Summary}

Data characterization techniques extract meaningful dataset properties that the downstream machine learning tasks and applications could use to improve performance. We hypothesize that the clustering indices computed from dataset clustering represent dataset characteristics concerning a specific clustering method. Clustering algorithms make different clustering assumptions for grouping the data points into neighborhoods.  Clustering indices are quality measures for validating the clusters induced by a clustering algorithm. When we use clustering indices to measure the performance of such clustering algorithms, they inherently capture different properties of the datasets. When a clustering index is independent of any external information, such as data labels, the index becomes an \emph{internal} index, or \emph{quality} index \cite{clustercrit}. On the contrary, when the clustering index uses data point labels, it becomes an \emph{external} index.

\begin{table}[h]
    \label{notations}
    \centering
    \begin{tabular}{c|cl}
    \toprule
    \textbf{Symbol\ \ } & \ \ \ \  & \textbf{Definition} \\
    \midrule
      $D_i$ & & Binary classification dataset \\
      $n, p$ & & Number of data points and the data dimensionality of a dataset \\
      $\langle X_i, y_i\rangle$ & & A tuple of data instance vector and the respective binary label \\
      $\langle\mathbf{X}, \mathbf{y}\rangle$ & & A tuple of data instance and the respective labels in matrix form \\
      $\mathcal{D}, N$ & & Set of binary class datasets and its cardinality \\
      $C_i, \mathcal{C}$ & & Model class and the set of model classes\\
      $I_i, \mathcal{I}$ & & Clustering index and the set of clustering indices \\
      $\mathbf{I},\hat{\mathbf{I}}$ & & Instance vector and matrix forms of the clustering indices\\
      $\mathcal{Q}(D;C)$ & & Function to estimate the max $F_1$ for a given model class $C$ \\
      $O$ & & Individual $F_1\in[0,1]$ for model class $C$ computed by $\mathcal{Q}(D;C)$ \\
      $\mathbf{O},\hat{\mathbf{O}}$ & & Vector and matrix forms of $F_1$ scores for all the model classes \\
      $A, \mathcal{A}$ & & Clustering method and the set of clustering methods \\
      $\mathcal{F}_i(D;A)$ & & Function to map a dataset D to an index $I_i$ using the method $A$\\
      $\mathbb{F}$ & & Set of mapping functions \\
      $t, m$ & & Number of clustering indices and model classes \\
      $B_{ij}, \mathcal{B}_i$ & & Subsample and the set of subsamples drawn from a dataset $D_i$ \\
      $h, b$ & & Size and Count of subsamples \\
      $R_i, \mathcal{R}$ & & Regressor to estimate the expected $F_1$ of a model class $C_i$ and its set\\
  
      \bottomrule
     \end{tabular}
     \vspace{5pt}
     \caption{Notations}
     \label{tab:notations}
\end{table}

Table \ref{tab:notations} lists the notations used for representing different entities. Given a binary labeled dataset $D = \{\langle X_{i}, y_{i}\rangle \}_{i=1}^{n}$, where the data instance vector $X_{i}~\in~\mathbb{R}^p$ and the binary class label $y_{i} \in \{{-1},1\}$, the objective of our automatic model selection system is to determine the best model class $C_{best}$ from a set of model classes $\mathcal{C} = \{C_{1}, C_{2}, \cdots, C_{m}\}$ that provides the best classification performance. We hypothesize that the clustering indices representing the characteristics of a dataset~$D$ shall strongly correlate with the expected classification performance of a model class $C_i\in\mathcal{C}$ for the dataset $D$.

Let $\mathcal{I}=\{I_1, I_2,\cdots, I_t\}$ be the selected clustering indices containing internal and external measures. Let $\mathbb{F}=\{\mathcal{F}_1,\mathcal{F}_2,\cdots, \mathcal{F}_t\}$ be the set of functions that map a given dataset $D$ to a clustering index $I$ defined as $\mathcal{F}_j(D;A):D\rightarrow I_j$, where the data instance matrix $\mathbf{X}$ of $D=\langle \mathbf{X}, \mathbf{y}\rangle$ transforms to a scalar cluster index value $I_j$. Each function $\mathcal{F}(D; A)$ represents running a clustering algorithm $A$ on the dataset $D$, followed by extracting several clustering indices $I\in\mathcal{I}$.   The function $\mathbb{F}(D; A)$ represents processing the dataset $D$ independently by all the functions $\mathcal{F}_j\in\mathbb{F}$ as a \emph{Multiple-Instruction Single Data (MISD)} operation.  
\begin{equation}
\label{eq:generate_features}
\mathbb{F}(D;A)\equiv\left[\mathcal{F}_1(D;A), \mathcal{F}_2(D;A),\ldots,\mathcal{F}_t(D;A)\right]^T
\end{equation}
Let the dataset transformation to the cluster indices feature space be defined as $\mathbb{F}(D;A):D\rightarrow\mathbf{I}$, where $\mathbf{I}=\left[I_1,I_2,\ldots,I_t\right]^T$. Let the average $F_1$ score be the performance metric for evaluating the model-fitness of a model class $C_i$. Let $R$ be a regression task that learns the mapping between the clustering indices and the expected classification performance of the model class ${C}_i$ defined as $R(\mathbf{I};{C}_i):\mathbf{I}\rightarrow[0,1]$. We define individual regressors $R_i$ for each model class $C_i$ as a set $\mathcal{R}=\{R_1,R_2,\ldots,R_m\}$. Now, the objective of the automatic model selection method is to find the best performing model class $C_{best}\in\mathcal{C}$ for a dataset $D$ based on the maximum output from each of the regressors in $\mathcal{R}$.
\begin{equation}
    i^*=\arg \max_{1\le i\le m} R_i(\mathbb{F}(D;A))
\end{equation}
\begin{equation}
    C_{best} = C_{i^*}
\end{equation}



Training a regressor $R_i$ for a model class $C_i$ requires several samples of the form $\langle \mathbf{I}, O\rangle$, where $\mathbf{I}=\mathbb{F}(D;A)$, and $O$ being the maximum classification performance score achievable (for instance, $F_1$ metric) for the tuned model class $C_i$ computed by the function $\mathcal{Q}(D; C_i)$.  

We understand that the cluster indices feature vector $\mathbf{I}$ is computed for each dataset $D$. Assuming we have a collection of datasets $\mathcal{D}=\{D_1, D_2,\ldots, D_N\}$ for training the regressor $R_i$, if we consider each dataset as a single instance vector of clustering indices, the number of training samples gets limited by the size of the dataset collection $\mathcal{D}$. It becomes hard to train the regressor model due to the shortage of training samples. To overcome the data shortage problem, we train the regression functions $R_i\in \mathcal{R}$ using the dataset subsamples instead of the full dataset. In this process, every dataset $D_i$ undergoes random subsampling with replacement to generate several subsamples, say $b$ subsamples of constant size $h$ as $\mathcal{B}_i = \{B_{i1},B_{i2},\cdots B_{ib}\}$, where $B_{ij} = \{d\ |\ d\sim D_i\}_{k=1}^h\text,\ ||B_{ij}||=h$. 

An advantage of using subsamples instead of the full dataset is generating more variability in the datasets used for training the regressors, making it robust to the dataset variance.  Another advantage is the ease of generating clustering indices from subsamples compared to working with large datasets in a single shot. In the \emph{single shot} mode, we run the clustering algorithms on the full dataset population to estimate the clustering indices.  Usually, running the clustering algorithms on the dataset samples (\emph{subsamples} mode) is significantly faster than running on the full population (\emph{single-shot} mode).
  
 

\subsection{Clustering Indices of Subsamples Vs. Population}\label{Whysubsamples}

We use the stratified random sampling method to create subsamples of a dataset. In this section, we analyze how the subsamples represent the characteristics of the data population. We attempt to establish that the cluster indices estimated for a whole population are similar to that of the subsamples by visualizing the cluster indices in a lower-dimensional space through \emph{t-SNE}~\cite{tsne} based visualization. 

\begin{figure}[t]
\centering
    \includegraphics[width=\textwidth,scale=0.4]{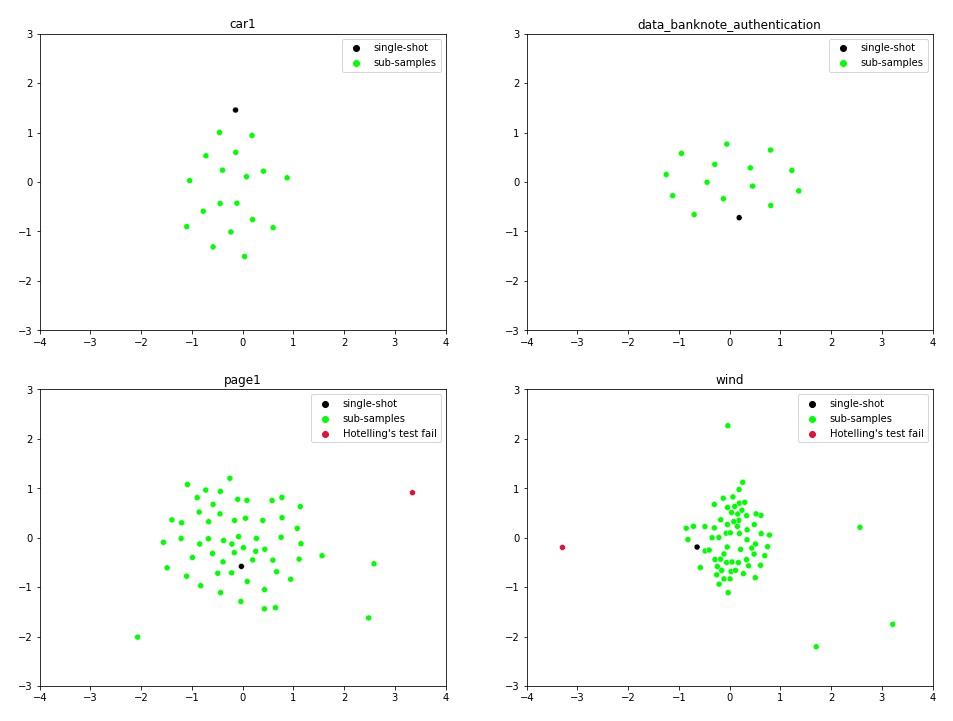}
    \caption{Low dimensional t-SNE visualization of the clustering indices estimated for the whole dataset population and dataset subsamples. The \textbf{BLACK} points represent the cluster indices of the whole dataset population, whereas the \textcolor{green}{\textbf{GREEN}} points represent cluster indices of the dataset subsamples. The \textcolor{red}{\textbf{RED}} points represent the subsamples that have failed Hotelling's $T^2$ test.}
    \label{fig:1shot_v_subsample}
\end{figure}

Figure \ref{fig:1shot_v_subsample} represents the lower-dimensional representation of the dataset's cluster indices for the data population and subsamples. We observe from the figure that the subsample cluster indices are found near and around the whole population cluster indices in most cases. The closeness of the cluster indices implies that the subsamples are indeed representative of the dataset population collectively.  

We use \emph{Hotelling's two-sample $T^2$ test} \cite{hotellings}, a multivariate extension of two-sample \textit{t}-test for checking if a subsample is representative of the whole dataset population in terms of the respective dataset predictor variables.  Figure~\ref{fig:1shot_v_subsample} shows the subsamples that fail the Hotelling's $T^2$ test (shown as RED dots) along with the passing subsamples (shown as GREEN dots). The subsamples that pass Hotelling's test appear closer to the whole dataset population shown in BLACK color.  A few subsamples appear relatively far from the entire population, but they seem to fail in Hotelling's test appropriately. This observation confirms the agreement of Hotelling's test response to the t-SNE visualization of the clustering indices.

One possible reason for some subsamples to go away from the whole population cluster indices is the randomness in sampling.  The deviant subsamples are beneficial in the training phase as our dataset construction pipeline described in Section \ref{data_construction} considers each subsample as an independent dataset. Therefore, the far-away subsamples offer new dataset variance to our regressor $\mathcal{R}$ training that, in turn, should help the regressors to achieve better generalization over unseen datasets.

On the other hand, the deviant subsamples are problematic in the recommendation pipeline, as they might skew the estimated classification performance of the model classes in $\mathcal{C}$. Removing such deviant subsamples from the experiment requires the whole population cluster indices that might not be available at the testing time.  To overcome this limitation, we propose to remove subsamples that fail the Hotelling's $T^{2}$ test from the recommendation pipeline described in Section \ref{Testing_phase} to limit the skew due to random subsampling. 

\subsection{Automatic Model Selection System Architecture}\label{MSsystem}
 \begin{figure}[t]
     \centering
     \includegraphics[width=\textwidth]{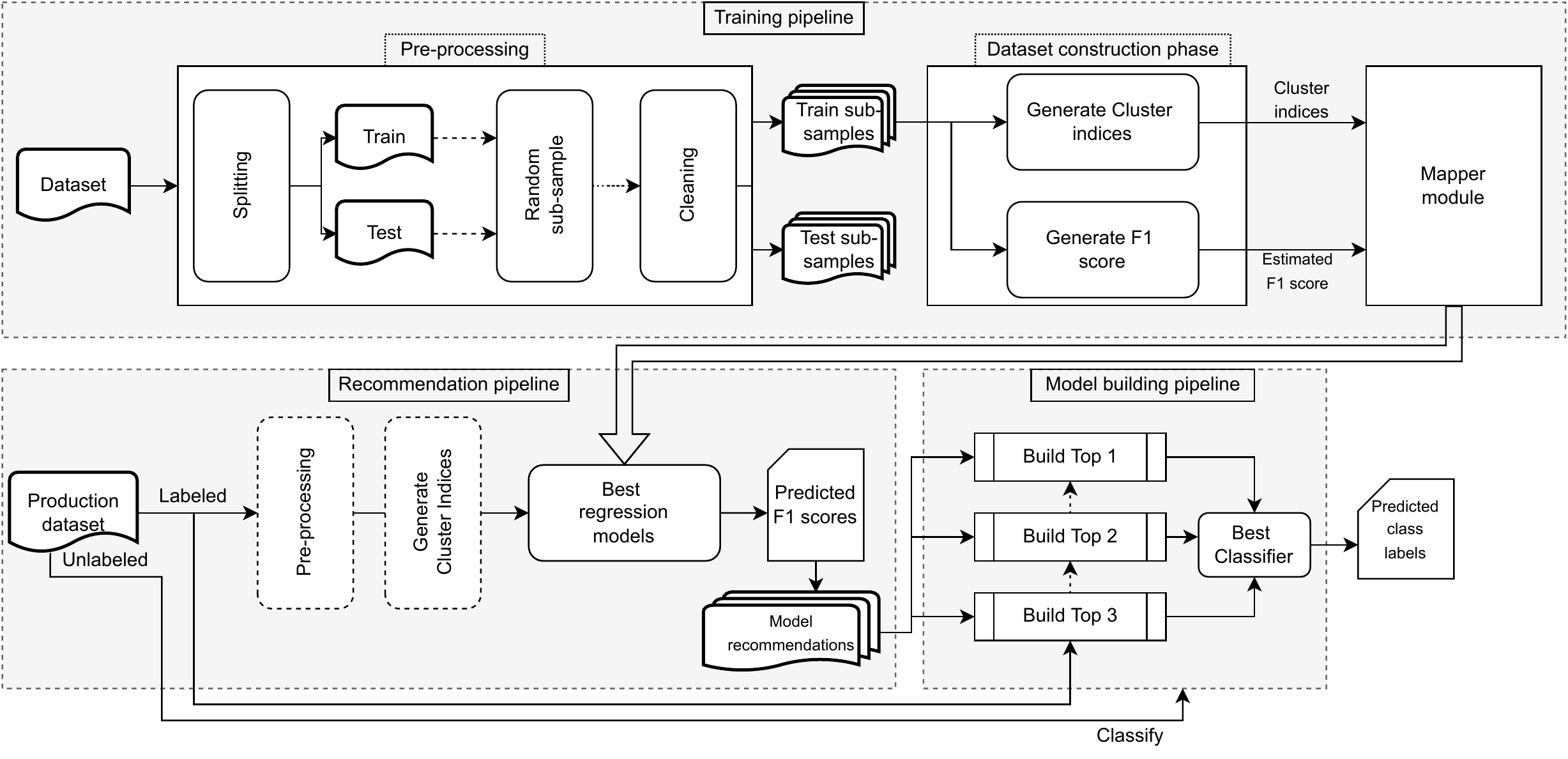}
     \caption{Architecture of the Automatic Model Selection System}
     \label{fig:MSsystem_fig}
 \end{figure}

The Automatic Model Selection system consists of two independent pipelines, one for training the underlying regression models and another pipeline for recommending the top-performing model classes for a given dataset. Figure \ref{fig:MSsystem_fig} illustrates the overall architecture of the proposed automatic model selection system.  The upper section is the \emph{Training} pipeline, and the bottom is the model \emph{Recommendation} pipeline. The \emph{Mapper} module of the training pipeline contains the regressors for each model class that learn the mapping between clustering indices and model fitness.  The model recommendation pipeline uses the learned regressors (Mapper modules) to predict the expected model fitness in terms of $F_1$ score (but not limited to) for each model class. The model classes that score the \emph{top3} $F_1$ scores become the best-fit classifier candidates for the given production dataset. We pick the best-performing classifier among the recommended top models through cross-validation to predict the labels of the unlabeled data. 



\subsection{Training Pipeline}
The training phase has three subparts, namely:- A. \textbf{\textit{Preprocessing}}, B) \textbf{\textit{Data Construction}}, and C) \textbf{\textit{Mappers as Regression models}} shown as the \emph{Training} pipeline enclosed in the top dotted rectangle region in Figure \ref{fig:MSsystem_fig}.

\paragraph{\textit{Preprocessing:}}\label{preprocess}

Data preprocessing involves multiple sub-tasks to transform a single dataset $D_i$ into a set $\mathcal{B}_i$ of several subsamples generated by stratified random sampling with replacement. The dataset $D_i$ divides into 70:30 training-validation partitions for cross-validating the regressor model training. The training and validation partitions undergo random sampling independently to generate the respective subsamples. The constructed subsamples \{$B_{ij}\}_{j=1}^b$ from a dataset $D_i$, undergo a cleansing process involving scaling and standardization. At the end of the preprocessing stage, we have a set $\mathcal{B}_i=\{B_{i1},B_{i2},\ldots,B_{ib}\}$ of several randomly sampled subsamples from both the training and validation partitions of the dataset $D_i$. The training process uses the training subsamples for learning and tuning the regressor functions $\mathcal{R}$ through cross-validation. We use the validation partitions for reporting the performance through $R^2$ score.
 
\begin{table}[ht]
\begin{tabularx}{\textwidth}{>{\hsize=.35\hsize}X|X} 
\hline
\textbf{Model Class} & \textbf{Hyper-parameters} \\
\hline
\vspace{0pt}
Logistic Regression &
\vspace{-10pt}
\begin{itemize}
    \item Penalty: Elasticnet
\end{itemize}
\\

\hline
\vspace{5pt}
Decision Tree &
\vspace{-15pt}
 \begin{itemize}
    \item Maxdepth: Tree grows until the leaves are pure, followed by backward-pruning for optimal depth
    \item Criterion: Gini index
\end{itemize}
\\
\hline
\vspace{10pt}
Regression Forest &
\vspace{-15pt}
 \begin{itemize} \item \#Trees: 100
\item Criterion: Gini index
\item Maxdepth: Tree grows until the leaves are pure, followed by backward-pruning for optimal depth
\end{itemize} \\
\hline
\vspace{5pt}
SVM & 
\vspace{-15pt}
\begin{itemize}
    \item Kernel: RBF
    \item C (Penalty): Grid search over \{0.01, 0.1, 1, 10, 100\}
    \item $\gamma$ (bandwidth) is set to $\frac{1}{p\times\sigma}$
    \begin{itemize}
        \item $\sigma$ is the variance of the flattened feature matrix $\mathbf{X}$
        \item $p$ is the data dimensionality
    \end{itemize}
\end{itemize} \\ 
\hline
\vspace{0pt}
K-NN & 
\vspace{-15pt}
\begin{itemize}
    \item \#Neighbors: Based on the Imbalance ratio (ex: K=R, if the ratio is 1:R)
\end{itemize} \\
\hline
\vspace{-5pt}
XGBoost & 
\vspace{-15pt}
\begin{itemize}
    \item Maxdepth: Grid search over \{1, 2, 4, 8, 16, 32, 64\}
\end{itemize} \\
\hline
\end{tabularx}
\vspace{2pt}
\caption{Hyper-parameters for tuning the classifiers while estimating the model-fitness}
\label{tab:classifier-hyperparameters}
\end{table}

\paragraph{\textit{Estimating model-fitness:}} \label{para:estimating-model-fitness}
When applied to a dataset, the model-fitness score indicates what to expect as the classification performance from a model class. We set up the function $\mathcal{Q}(B;C)$ to measure the model-fitness of a classifier model class $C$ for the dataset $B$. We measure the model-fitness by estimating the maximum achievable classification performance measured using $F_1$ metric (but not limited to) by building tuned classifiers for the given dataset. Table~\ref{tab:classifier-hyperparameters} lists the tuning parameters we use for each model class to achieve optimal performance. We tune the classifiers for each of the given datasets to find the respective maximum achievable $F_1$ score and use the estimate as a surrogate measure for model-fitness. Our hyper-parameters list is not exhaustive but only indicative of the need for optimizing the model performance. We do this exercise through cross-validation to avoid an overfitting scenario that potentially skews the model-fitness score.

\paragraph{\textit{Data construction}:} \label{data_construction}
Given a set of subsamples $\mathcal{B}_i=\{B_{i1},B_{i2},\ldots,B_{ib}\}$ for training and validation drawn from each dataset $D_i\in\mathcal{D}$, the objective of the data construction phase is to generate a set of tuples  $\{\langle \mathbf{I_{ij},O_{ij}}\rangle\}_{j=1}^b$ from all the subsamples in $\mathcal{B}_i$, where $\mathbf{I_{ij}}=\mathbb{F}(B_{ij})$ as per Eq. \ref{eq:generate_features} and $\mathbf{O_{ij}}$ is the model-fitness score for a dataset $B_{ij}$ for every model class in $\mathcal{C}$ given by:

\begin{equation}
    \mathbf{O_{ij}}\leftarrow\left[ \mathcal{Q}(B_{ij};C_1), \mathcal{Q}(B_{ij};C_2),\ldots, \mathcal{Q}(B_{ij};C_m)\right]^T
\end{equation}





At the end of the data construction phase, we get a matrix of clustering indices feature vectors $\hat{\mathbf{I}}_i=[\mathbf{I}_{i1}, \mathbf{I}_{i2},\ldots,\mathbf{I}_{ib}]^T$ generated for every subsample from the set $\mathcal{B}_i$ from dataset $D_i$ with the corresponding matrix of model-fitness scores $\hat{\mathbf{O}}_i=[\mathbf{O}_{i1}, \mathbf{O}_{i2},\ldots,\mathbf{O}_{ib}]^T$ estimated for each model class in $\mathcal{C}$.  Then, we combine the data generated for individual training datasets $D_i\in\mathcal{D}$ into a jumbo dataset $\langle\hat{\mathbf{I}},\hat{\mathbf{O}}\rangle$, such that:
\begin{equation}
    \hat{\mathbf{I}}\leftarrow\left[\hat{\mathbf{I}}_1,\hat{\mathbf{I}}_2,\ldots,\hat{\mathbf{I}}_N\right]^T
\end{equation}
\begin{equation}
    \hat{\mathbf{O}}\leftarrow\left[\hat{\mathbf{O}}_1,\hat{\mathbf{O}}_2,\ldots,\hat{\mathbf{O}}_N\right]^T
\end{equation}

\paragraph{\textit{Mapper for model selection:}} \label{AutoMS_mapper}
In the \emph{Mapper} phase, we learn a multiple regression function $\mathcal{R}:\hat{\mathbf{I}}\rightarrow\hat{\mathbf{O}}$ using the dataset $\langle\hat{\mathbf{I}},\hat{\mathbf{O}}\rangle$ generated from the \emph{Data Construction} stage. We tune the hyper-parameter of the multiple regressors using the evaluation set through cross-validation. Alternately, instead of multiple regressors $\mathcal{R}$, we can also build individual regressors $R\in\mathcal{R}$ per model class $C\in\mathcal{C}$ as $R_k:\hat{\mathbf{I}}\rightarrow\hat{\mathbf{O}}^{(k)}$, where $\hat{\mathbf{O}}^{(k)}$ is the $k^{th}$ column-vector of the matrix $\hat{\mathbf{O}}$.
The mapper module constitutes the resulting set of regressors $\mathcal{R} = \{R_1, R_2,\ldots, R_m\}$, which we use to predict the expected classification performance of different model classes $C_k\in\mathcal{C}$ for a given test dataset $D'$. 

During prediction, the mapper estimates the expected classification performance of a dataset from its clustering indices features.  The expected classification performance is what an optimized/tuned classifier may achieve for a given dataset.  We assume that the representation of the dataset samples in the clustering indices space follows the \emph{i.i.d} assumption.  This means that the parameter setting required for achieving higher performance for the training sample shall be similar to that of the validation sample. As the mapper learns to map the clustering indices to the tuned classifier performance during training, we expect the mapper to predict the closest estimate of optimized performance during validation. 

\subsection{Recommendation Pipeline}\label{Testing_phase}

The automatic model selection system's recommendation pipeline is a simple process of invoking the tuned regressor models $\mathcal{R}$ for predicting the expected model-fitness measured in terms of expected classification performance for a given test dataset $D'$. The test dataset undergoes the same data transformation and cleansing stages as the training pipeline for consistency. 
\begin{equation}
    \mathbf{I}'\leftarrow\mathbb{F}(D')
\end{equation}

The transformed data is input to each regressor function $R\in\mathcal{R}$ to predict the expected classification performance (model-fitness) score for all the model classes $C\in\mathcal{C}$. 
\begin{equation}
    \mathbf{O}'\leftarrow\mathcal{R}(\mathbf{I}')
\end{equation}

We recommend the best model class $C_{best}$ that scores the highest model-fitness score for the given test dataset $D'$.

\begin{equation}
\label{argmax1}
    i^* \leftarrow \arg \max_{\forall R_i\in\mathcal{R}} R_i(\mathbf{I}')\equiv
    \arg \max_{1\le i\le m} \mathbf{O}'_i
\end{equation}
\begin{equation}
\label{bestc1}
    C_{best} \leftarrow C_{i^*}
\end{equation}

Alternately, the prediction for the test dataset is also runnable using the data set subsamples, where we generate several subsamples $\mathcal{B}'$ by random sampling with replacement from the input test dataset $D'$ as $\mathcal{B}'=\{B'_1,B'_2,\ldots,B'_b\}$, where $B'_i = \{d\ |\ d\sim D'\}_{k=1}^h,\forall B'_i\in\mathcal{B}'$. We then transform the subsamples to the clustering index feature space.

\begin{equation}
    \mathbf{I}'_j\leftarrow\left[\mathcal{F}_1(B'_j),\mathcal{F}_2(B'_j),\ldots,\mathcal{F}_t(B'_j)\right]^T, \forall B'_j\in\mathcal{B}'
\end{equation}

We input these vectors of cluster indices to the regressors $\mathcal{R}$ to make predictions of expected model-fitness scores for each of the model classes $C\in\mathcal{C}$.

\begin{equation}
    \mathbf{O}'_j\leftarrow\mathcal{R}(\mathbf{I}'_j), 1\le j\le b
\end{equation}

We compute the expected model-fitness scores for all the model classes for the dataset $D'$ by averaging the estimated fitness scores for each subsample $B'\in\mathcal{B}'$.
\begin{equation}
    \label{expf1}
    \mathbf{O}'\leftarrow\frac{1}{b}\sum_{j=1}^b \mathbf{O}'_j
\end{equation}

With the availability of the estimated $F_1$ scores, the expected classification performance per model class from Equation \ref{expf1}, we use Equations \ref{argmax1} and \ref{bestc1} to recommend the best model class $C_{best}$ for the test dataset $D'$.




\begin{table}
    \centering
    \scalebox{1}{
    \begin{tabular}{c|c}
    \toprule
        \textbf{Internal Indices\ \ } & \textbf{\ \ External Indices}\\
    \midrule
        Between-cluster scatter & Entropy \\
        Banfeld-Raftery & Purity\\ 
        Ball-Hall & Recall \\ 
        PBM  & Folkes-Mallows\\
        Det-Ratio & Rogers-Tanimoto\\
        Log-Det-Ratio & $F_1$ \\ 
        Ksq-DetW & Kulczynski\\ 
        Score  & Norm-Mutual-Info\\
        Silhouette & Sokal-Sneath-1\\
        Log-SS-Ratio & Rand \\ 
        C-index & Hubert $\Gamma$\\
        Dunn & Homogeneity \\
        Ray-Turi & Completeness\\
        Calinski-Harabasz & V-Measure\\
        Trace-WiB & Jaccard\\ 
        Davies-Bouldin & Adj-Rand \\
        Within-cluster scatter & Phi \\
        & McNemar \\
        & Russel-Rao \\
        & Precision \\
        & Weighted-$F_1$ \\
        &Sokal-Sneath-2\\
        & Adj-Mutual-Info\\
    \bottomrule
    \end{tabular}
    }
    \vspace{5pt}
    \caption{Set of internal and external clustering indices $\mathcal{I}$}
    \label{tab:indices_list}
\end{table}

\begin{table}
\centering
\begin{tabular}{cc|cc}
\toprule
\textbf{Model Class $\mathcal{C}$} & \textbf{Family}   & \textbf{Clustering} & \textbf{Family} \\ \midrule
Decision Tree       & Divisive          & K-Means++              & Convex          \\
Logistic Regression & Linear            & \ \ Agglomerative       & \ \ Hierarchical     \\
SVM                 & Large Margin      & Spectral            & Kernel, Non-convex          \\
KNN                 & Lazy Learner      & HDBSCAN             & Density, Non-convex         \\
Random Forest       & Bagging Ensemble &                     &                 \\
XGBoost             & Boosting Ensemble\ \  &                     &                 \\ \bottomrule
\end{tabular}
\vspace{5pt}
\caption{List of classifiers $C_i\in\mathcal{C}$ representing different model classes and the set of clustering methods $\mathcal{A}$ to generate clustering indices $\mathcal{I}$.}
\label{tab:methods}
\end{table}

\subsection{Configuration}\label{config}

The Automatic Model Section system requires configuration of the clustering indices set $\mathcal{I}$ and the list of representative model classes $\mathcal{C}$. Table \ref{tab:indices_list} lists the set of clustering indices \cite{cluster_crit} that we configure the system to use for representing the dataset characteristics. The dataset clustering assumptions greatly influence the clustering indices. To make the clustering assumptions comprehensive, we configure multiple clustering algorithms representing different clustering assumptions on the dataset. Table \ref{tab:methods} lists different clustering algorithms $A\in\mathcal{A}$ that we use to generate the respective clustering indices. We concatenate the clustering indices generated by these \emph{4 (four)} clustering methods to form a broader clustering index feature space of $4t$ dimensions. 
\begin{equation}
    \mathcal{I} = \left[\mathcal{I}_{A_1:kmeans}, \mathcal{I}_{A_2:agglomerative}, \mathcal{I}_{A_3:spectral}, \mathcal{I}_{A_4:hdbscan}\right]
\end{equation}
\begin{equation}
    \mathcal{I} = \left[\underbrace{I_1,I_2,\ldots,I_t}_{A_1:kmeans},\underbrace{I_{t+1},I_{t+2},\ldots,I_{2t}}_{A_2:agglomerative},\underbrace{I_{2t+1},I_{2t+2},\ldots,I_{3t}}_{A_3:spectral},\underbrace{I_{3t+1},I_{3t+2},\ldots,I_{4t}}_{A_4:hdbscan}\right]
\end{equation}

The transformation function set becomes $\mathbb{F}=\{\mathcal{F}_1,\mathcal{F}_2,\ldots, F_{4t}\}$ to cover the clustering indices from four different families of dataset clustering assumptions.
We aim to use as many clustering indices as possible to build the dataset's feature space representation. By scaling up the \emph{40 (forty)} dimensional clustering indices features from Table \ref{tab:indices_list} with \emph{four} different clustering algorithms from Table \ref{tab:methods}, we get a total of $4\times40=160$ clustering indices features to represent the dataset characteristics. The mapper modules use an appropriate subset of the features while learning the association between clustering indices and the expected classification performance of a model class.
\begin{table}
    \centering
    \scalebox{1}{
    \begin{tabular}{ccc|ccc}
    \toprule
    \textbf{Dataset} & \textbf{Dims} & \textbf{Count} & \textbf{Dataset} & \textbf{Dims} & \textbf{Count} \\ \midrule
    analcatdata\_halloffame & 17 & 1340 & 2d planes & 10 & 40767\\ 
abalone & 8 & 4176 & contraceptive1\_2 & 9 & 1473 \\
balloon & 2 & 2000 & banana & 2 & 5299 \\
 bank32nh & 32 & 8192 & car1 & 6 & 1728 \\ 
 chess & 36 & 3196 & churn & 20 & 4999 \\ 
 cmc21 & 10 & 1473 & coil2000 & 85 & 9822 \\ 
 connect4 & 43 & 67557 & contraceptive1\_1 & 9 & 1473 \\ 
 adult & 14 & 48842 & contraceptive1\_3 & 9 & 1473 \\ 
 data\_banknote\_auth & 5 & 1372 &  cpu\_small & 12 & 8192\\ 
  flare & 11 & 1066 & default\_of\_credit & 16 & 20000 \\ 
 credit\_card\_clients & 20 & 1000 & elevators & 18 & 16599 \\ 
 fried & 11 & 40768 &  hill valley & 100 & 1211 \\ 
 HTRU\_2 & 9 & 17898 & image desert & 136 & 2000 \\ 
 kdd\_japaneseVowels & 15 & 9961 & kc1 & 21 & 2108\\ 
 kin8nm & 9 & 8190 &  kr-vs-kp & 36 & 3196 \\ 
 magic04 & 10 & 19020& mushroom & 22 & 5644 \\ 
 musk2 & 170 & 6598 & mv & 10 & 40768 \\ 
 optdigits & 64 & 5620& page1 & 10 & 5472 \\ 
 pc1 & 21 & 1107 & pendgits & 16 & 10991 \\ 
 phoneme & 6 & 5404 &  pizzacutter3 & 38 & 1043 \\ 
 pollen & 6 & 3848& pol & 48 & 15000 \\ 
 puma32H & 33 & 8192 & qsar\_biodeg & 42 & 1055 \\ 
 quake & 4 & 2178 & ring & 20 & 7400 \\ 
 satimage1 & 36 & 6435 & spambase & 57 & 4597 \\ 
 splice1 & 60 & 3190& splice2 & 60 & 3190 \\ 
 splice3 & 60 & 3190& steel\_plates & 33 & 1940 \\ 
 tic-tac-toe & 10 & 957 & titanic & 4 & 2200 \\ 
 tokyo & 45 & 959 & twonorm & 20 & 7399 \\ 
 vehicle31 & 18 & 845 &  visualizing\_soil & 5 & 8641 \\ 
 wind & 15 & 6574 & waveform\_5000 & 40 & 5000  \\ \bottomrule
\end{tabular}}
\vspace{5pt}
\caption{List of binary class datasets used for training and validating the proposed model selection }
\label{tab:datasets}
\end{table}

\subsection{Classification Modeling As A Service}
\label{sec:CMAAS}
In an Automated ML setting, a user uploads a production dataset containing labeled and unlabeled parts to the service endpoint.  The user then expects label predictions for the unlabeled part of the dataset without thinking about the underlying machine learning pipeline. As an enterprise extension to the Automatic Model Selection system, we expose the underlying data classification modeling as a service (an Automated ML SaaS offering) by abstracting the model selection, tuning, and training processes on a \emph{production} dataset.    Here, we rank the model classes by their estimated model fitness score and pick the best performing model among \emph{top3} models ($C_A, C_B, C_C\in\mathcal{C}$). We use the given \emph{labeled} portion of the production dataset $\langle\mathbf{X},\mathbf{y}\rangle$ to cross-validate the \emph{top3} models with the best parameter settings. Using the best-performing model amongst the \emph{top3}, we expose an API for predicting the class labels for the \emph{unlabeled} portion of the production dataset.  When the user invokes the API with the unlabeled data instance, the best-performing model among \emph{top3} classifiers outputs the class label $\hat{y}$ for the data instance $X$ from the test dataset $D'$.


We describe the detailed validation of the end-to-end Automated ML system in Section \ref{validating_saas} by comparing it against a few of the popular commercial and noncommercial Automated ML solutions.

\section{Experimental setup} \label{expsetup}
In this section, we test our \textbf{C}lustering \textbf{I}ndices based \textbf{A}utomatic classification \textbf{M}odel \textbf{S}election (\emph{CIAMS}) hypothesis by cross-validating with several classification datasets collected from multiple public domain sources. 
We set up the experiment by choosing \emph{6 (six)} different classification model families listed in Table~\ref{tab:methods} representing various model classes. 


\subsection{Training Phase}
We use \emph{sixty (60)} binary class datasets in Table \ref{tab:datasets} to build our Automatic Model Selection system.  We divide the datasets into \emph{train} and \emph{test} partitions. We use the \emph{train} partition to build and tune our model selection system and the \emph{test} partition to report the performance results.  We divide \emph{train} partition further into \emph{trainset} and \emph{evalset} partitions in a $k$-fold cross-validation setting to tune our model selection system.  Each dataset from \emph{trainset} and \emph{evalset} partitions undergo sub-sampling independently. This ensures that we do not reuse any training data points during validation.  Dataset subsampling increases the number of data points for training the mapper module (regressors) and exposes the underlying regressor models to more data variances.  In total, we create \emph{11,190} subsamples from all the $60$ datasets.   


\subsection{Evaluation Phase}
\label{evalphase}
We evaluate the performance of our Automatic Model Selection system using two modes, such as A) \emph{subsamples} mode and B) \emph{single-shot} mode.  The \emph{subsamples} mode mimics the dataset preparation strategy of the Training Phase, where we use only the subsamples of the datasets in the \emph{test} partition for running the performance evaluation. Each dataset from the \emph{test} partition gives out several test data points constructed from several subsamples drawn from it. 

We set up cross-validation on 60 datasets using their respective subsamples.  
Traditionally, the sizes of the training and evaluation splits remain constant across all the cross-validation folds.  In our setting, the split sizes vary proportionally to the population sizes of the datasets picked under the training and evaluation splits.  We explain the proportionality in Section \ref{sec:hyperparameters}. We aim to construct the folds without any dataset spilling across folds.  A fold may contain one or more datasets, but a dataset restricts to only one fold.

An ideal configuration to run the evaluation is \emph{Leave-One-(dataset)-Out}.  In~our setting, the equivalent is running a $60$-fold cross-validation. We observe that 60-fold cross-validation on the subsamples of 60 datasets is time-consuming.  To speed up, we repeat the 6-fold cross-validation exercise several times to approximate the results of 60-fold. Also, repeating the cross-validation several times allows us to test our method with different combinations of datasets in the training and testing folds.

On the contrary, we evaluate the \emph{single-shot} mode through \emph{Leave-One-Out} $60$-fold cross-validation. The single-shot mode uses the entire dataset from the \emph{test} partition as one test record.  The \emph{single-shot} mode evaluation helps assess the system's usefulness in enterprise deployment settings. An advantage of the \emph{subsamples} mode is the ability to work with large datasets, as the full dataset \emph{single-shot} approach might become resource-intensive for generating the clustering indices.


\subsection{Hyper Parameters}
\label{sec:hyperparameters}
A few hyperparameters control the system's behavior, which we tune by trial and error.

\paragraph{\textit{Number of Clusters}:}The number of clusters is a critical parameter for most clustering algorithms. We set the \emph{number-of-clusters} parameter to $2$ for the clustering algorithms listed in Table \ref{tab:methods}. Initially, we allow a linear search for the number-of-cluster hyper-parameter. Our linear search experiment achieves the best results when the count is $2$. 

\paragraph{\textit{Subsamples}:} We set the subsample size as $h=500$ after doing a linear search with different subsample sizes when the size of the dataset $n$ is beyond \emph{2000} points.  We choose the following $h$-value for smaller datasets depending on the dataset size range.
\[  h \leftarrow \left\{
\begin{array}{lcl}
    100 & \text{    } & n\le 500\\
    300 & \text{    } & 500<n\le 2000\\
    500 & \text{    } & n>2000\\
\end{array} 
\right. \]
The number of subsamples $b$ is set proportional to the size of the dataset $n$. The average bootstrap subsample contains 63.2\% of the original observations and omits 36.8\% \cite{10.5555/2161980}. When the subsample size is $h$, we get $0.63h$ data points from the original dataset. The following equation helps us get the number of subsamples to draw from the dataset to ensure maximum coverage of dataset variances and more datasets for training our internal models.  The hyper-parameter $\alpha$ is the oversampling constant, which we set to $5$ based on trial and error.

\begin{equation*}
    b\leftarrow\alpha\times\left\lceil\frac{n}{0.63 h}\right\rceil
\end{equation*}

\subsection{Scaling Up}
Our present design of the \emph{CIAMS} system uses \emph{four (4)} clustering models on the datasets, \emph{six (6)} classification model classes, and \emph{sixty (60)} binary class datasets.  We can train the underlying mapper module with more datasets $\mathcal{D}$ seamlessly to make the regressors robust for handling dataset variances.  We can also add more model classes to $\mathcal{C}$ to increase the choices of classification methods for a given dataset. For comparison, the Azure AutoML\footnote{https://docs.microsoft.com/en-us/azure/machine-learning/concept-automated-ml} platform uses about 12~classifiers\footnote{https://docs.microsoft.com/en-us/azure/machine-learning/how-to-configure-auto-train} from 8 model classes. \emph{Neural Networks} and \emph{Bayesian methods} are the two extra model classes other than the model classes listed in $\mathcal{C}$.  Likewise, we can further expand the clustering indices feature space by adding other clustering methods to the set $\mathcal{A}$. The \emph{CIAMS} system is scalable and extensible by design. 

\section{Evaluation}\label{results}

In this section, we evaluate the \emph{CIAMS} system at different granularity levels in a bottom-up style to answer the following questions that may challenge our proposed scheme's usefulness.
\begin{enumerate}
    \item Is the dataset model-fitness a function of the dataset clustering indices for a model class?
    \item Does the \emph{Mapper} module implemented using \emph{Regression} models efficiently learn the relationship between the dataset clustering indices and the expected classification performance of a selected model class? 
    \item What is the correctness of the recommendation given by the \emph{CIAMS} system? 
    \item What is the performance of the end-to-end \emph{Classification Modeling as a Service} platform for enterprise deployments?
\end{enumerate}

\subsection{Mapper Module Evaluation} 

Firstly, we evaluate the performance of the \emph{mapper} module constructed using Regression models. Evaluating the regressors helps us understand the validity of our model selection hypothesis, which assumes that the clustering indices of a dataset strongly correlate with the expected classification performance of a chosen model class.  We evaluate our idea using the estimated $R^2$ metric for every regressor built for every classifier model class. We also study the usefulness of the regressors through $L_1$-norm, or \emph{Mean Absolute Error} (MAE) analysis between the predicted and actual dataset model-fitness measured using $F_1$ score. We run a small-sample statistical significance test on the $L_1$-norm estimates to check if the system contains the error margin within $\pm 10\%$.

\begin{table}
    \centering
    \begin{tabular}{c|c|c}
    \toprule
    \textbf{Model Class} & \emph{\ Single-shot\ } & \emph{\ Subsamples\ } \\
    $C_i\in\mathcal{C}$ & 60-fold & 6-fold \\
    \midrule
    {Decision Tree} & 0.83 & 0.81  \\
    {Random Forest} & 0.89 & 0.86  \\
    {Logistic Regression } &  0.87& 0.84  \\
    {K-Nearest Neighbor } & 0.80 &  0.80 \\
    {XGBoost } & 0.85 &  0.83 \\    
    {SVM} & 0.81 & 0.81   \\ 
    \midrule
    {Average} & 0.84 & 0.83 \\
    \bottomrule
    \end{tabular}
    \vspace{5pt}
    \caption{Performance of the tuned Regressor functions $R_i\in\mathcal{R}$ for each model class $C_i\in\mathcal{C}$ measured using $R^2$ score in \emph{Single-Shot} and \emph{Subsamples} modes.}
    \label{tab:regressor-r2-score}
\end{table}

\subsubsection{Regressor Performance}
We designate a regressor $R_i$ for every classifier $C_i$ from the set of model classes $\mathcal{C}$.  Each regressor function $R_i:\hat{\mathbf{I}}\rightarrow\hat{\mathbf{O}}^{(i)}$ learns a mapping between the cluster indices $\hat{\mathbf{I}}$ and the tuned classifier $C_i$ performance $\hat{\mathbf{O}}^{(i)}$ measured in $F_1$ metric for all the training datasets $\mathcal{D}$.  We experiment with several regression models such as SVR, Random Forest, XGBoost, k-NN, and Decision Tree and select XGBoost to be the best regression model to learn the mapper through $R^2$ analysis.   We train the regressor models using two configurations, namely:- A) \emph{single-shot}, and B) \emph{subsamples} mode as explained in Section~\ref{evalphase}.  In the \emph{single-shot} mode, we run \emph{Leave-One-(dataset)-Out} cross-validation, and in the \emph{subsamples} mode, we run $6$-fold cross-validation with \emph{six} repeats to report the performance of the regressor models measured using $R^2$ score. We tune the XGBoost regressor models for best parameters by cross-validating on the \emph{train} partition that gets split further into \emph{trainset} and \emph{evalset} partitions per fold. We repeat this exercise for all the classifiers in $\mathcal{C}$ to find the best XGBoost regression model parameters that maximize the $R^2$ score.  Table \ref{tab:regressor-r2-score} narrates the $R^2$ scores of the tuned regressors $R_i\in\mathcal{R}$; we build for every classifier $C_i\in\mathcal{C}$ in \emph{Single-Shot} and \emph{Subsamples} modes. We observe an average of 84\% $R^2$ score for both modes, which confirms the feasibility of learning a mapping between clustering indices and model-fitness.


\subsubsection{Prediction Correctness}
We validate the regressors by checking if the predicted classification performance is similar to a model class's expected classification performance on a dataset.  We measure the similarity between the predicted and the expected classification performance measured as the model-fitness score ($F_1$) using \emph{Mean Absolute Error (MAE)} or $L_1$-norm. 
\begin{equation}
    MAE \leftarrow \left\|F_1^{\;expected}-F_1^{\;predicted} \right\|
\end{equation}
For every model class $C_i$, we consider the prediction of the mapper module (regressor $R_i$) to be a \texttt{PASS} if the absolute difference between the predicted and the actual classification performance is within 10\% margin. 
\[  \texttt{Correctness} \leftarrow \left\{
\begin{array}{lcl}
    \texttt{\textbf{PASS}} & \text{    } & \emph{MAE}\le10\% \\
    \texttt{\textbf{FAIL}} & \text{    } & \emph{MAE}>10\%
\end{array} 
\right. \]

\begin{table}
    \centering
    \begin{tabular}{c|c|c|c|c}
    \toprule
    \multicolumn{1}{c}{\textbf{\textit{Configuration}}} & \multicolumn{2}{|c|}{\textit{Single-shot (60-fold)}} & \multicolumn{2}{c}{\textit{Subsamples (6-fold)}}\\ 
    \midrule
    \textbf{Model Class} & \textbf{\#\texttt{PASS}} & \textbf{MAE} & \textbf{\#\texttt{PASS}} & \textbf{MAE}\\
    \midrule
    {Decision Tree} &32&$0.14\pm 0.11$ &38&$0.09 \pm 0.05$\\
    {Random Forest} &22&$0.13\pm 0.09$  &46&$0.08 \pm 0.04$\\
    {Logistic Regression} &23&$0.17\pm 0.13$  &37&$0.10 \pm 0.07$\\
    {K-Nearest Neighbour} &22&$0.13\pm 0.09$  &31&$0.12 \pm 0.08$\\
    {XGBoost} & 29&$0.11\pm 0.08$  &46&$0.08 \pm 0.05$\\
    {Support Vector Machine} &23&$0.15\pm 0.11 $  &25&$0.13 \pm 0.08 $\\
    \midrule
    Average & 25 & & 37 & \\
    \bottomrule
    \end{tabular}
    \vspace{5pt}
    \caption{Mapper module performance in terms of Prediction Correctness using \emph{MAE}.}
    \label{tab:mae_test}
\end{table}

Table \ref{tab:mae_test} summarizes the \emph{MAE} and \texttt{PASS} performance of the mapper module. It is apparent from the table that the \emph{MAE} is contained within $10\%$ for the majority of the datasets.  In the \emph{Subsamples} mode, the average MAE is slightly off for \emph{KNN}, and \emph{SVM} because of higher variance in the predicted $F_1$ scores.  It is interesting to observe that the Mapping module achieves 10\% compliance for over $75\%$ of the datasets for the ensembling methods with lower variance. The large-margin model class seems to have trouble with the stability in performance across different subsamples. An average of $37$ datasets have responded well to satisfy the $10\%$ error margin constraint.  When the average \emph{MAE} for a model class is within the $10\%$ margin, we are accurate in our prediction for over $61\%$ datasets.   Ignoring the low-performing SVM classifier, the truncated average PASS performance stands at $40$, which is two-thirds of the datasets. The result of the subsampling mode is positively motivating to further this research avenue into achieving higher performance with better subsampling and model class tuning. On the other hand, in the \emph{Single-shot} mode, the average \emph{MAE} estimates are generally above the $10\%$ margin. Although the error margin is higher than $10\%$, the Mapper module makes accurate predictions for around $41\%$ of $60$ datasets.  It is obvious from Table \ref{tab:mae_test} that the \emph{Subsamples} mode is more appropriate than the \emph{Single-shot} mode for making predictions because the \emph{Mapper} module inherently uses only the \emph{subsamples} for training.

We perform the statistical significance test on the \emph{MAE} estimate to check if the mapper module is indeed restricting the \emph{MAE} within the \emph{10\%} margin.  We use the following version of the $t$ and $Z$-statistic \cite{bluman_2012} to perform the significance test. Suffixes $e$ and $p$ denotes the expected and predicted model-fitness ($F_1$) scores, $\Delta$ is the hypothesized difference between the population \emph{means} $\Bar{x}_e,\Bar{x}_p$ (we set $\Delta=0.1$).  

\begin{equation}
Z\text{ or }t\text{-statistic} \rightarrow \frac{\left\|\Bar{x}_e - \Bar{x}_p\right\| - \Delta} {\sqrt{\frac{\sigma^2_e}{n_e}+\frac{\sigma^2_p}{n_p}}}   
\end{equation}

We run $t$-test when the number of samples is $\le30$, and $Z$-test when the sample count is $>30$ at $p=0.05$ significance. Table \ref{tab:z-t - Test-6fold} summarizes the performance of the \emph{Mapper} module tested against \emph{60} datasets, where the second column lists the number of datasets where the \emph{MAE} between expected and predicted model-fitness ($F_1$) score is within the $10\%$ margin at the statistical significance level of $p=0.05$.  The $t / Z$-test confirms our regressors' ability to accurately predict the expected classification performance for at least $60\%$ of the datasets.  





\begin{table}[H]
    \centering
    \begin{tabular}{c|c}
    \toprule
    {Model Class} & \textbf{\ \#\texttt{PASS}\ } \\
    \midrule
    {Decision Tree} &37\\
    {Random Forest} &33\\
    {Logistic Regression} &41\\
    {K-Nearest Neighbour} &34\\
    {XGBoost} & 41\\
    {SVM} &32\\
    \midrule
    Average & 36\\
    \bottomrule
    \end{tabular}
    \vspace{5pt}
    \caption{Performance evaluation of the \emph{Mapper} module using two-sample $Z$-test \& $t$-test in the \emph{Subsamples} mode at $p=0.05$ significance}
    \label{tab:z-t - Test-6fold}
\end{table}

An average of \texttt{above 60\% PASS} datasets in Tables \ref{tab:mae_test} and \ref{tab:z-t - Test-6fold} is interesting because, in the subsamples mode, we test the ability of the mappers to generalize on 10 unseen datasets per cross-validation fold.  Out of 10 test-fold datasets, the mappers correctly predict the model-fitness for at least six datasets on average. A recall of 61\% demonstrates excellent promise in the technique and increases the motivation to improve this idea further to achieve even higher recall.
It is evident from the results in Tables \ref{tab:regressor-r2-score}, \ref{tab:mae_test}, and \ref{tab:z-t - Test-6fold} that the \emph{Mapper} module is efficient in learning the relationship between the clustering indices of a dataset and the expected classification performance of a model class.  The results also empirically prove the validity of our hypothesis that the dataset model fitness is indeed a function of the dataset clustering indices for a model class.

\subsection{Comparing Against Equivalent Methods}
\label{section:comparing-equivalent-methods}
We argue that the clustering indices features represent the data characteristics.  It is necessary to compare our approach against similar methods from the literature, such as Landmarking \cite{landmarker2}. Landmarking determines the location of a specific learning problem in the space of all learning problems by measuring the performance of some simple and efficient learning algorithms. Landmarking attempts to characterize the data properties by building simple classifier models. Similarly, classic statistical and information-theoretic features directly represent the data characteristics.  We compare the ability of the Landmarking and classic statistical, and information-theoretic features against clustering indices features to represent the data characteristics concerning a classification task.  

\begin{table}
\centering
\begin{tabular}{cl}
\toprule
\\
\multicolumn{2}{c}{\textbf{Classic Statistical Features}} \\ 
\\
\toprule
\multicolumn{1}{c|}{\textit{Basic Statistics}} & \begin{tabular}[c]{@{}l@{}}Number of points\\ Number of attributes\\ Number of classes\end{tabular} \\ \hline
\multicolumn{1}{c|}{\textit{Feature-based}} & \begin{tabular}[c]{@{}l@{}}F1- Maximum Fisher's discriminant ratio\\ F1v - Directional vector maximum Fisher’s discriminant ratio\\ F2 - Volume of the overlapping region\\ F3 - Maximum individual feature efficiency\\ F4 - Collective feature efficiency\end{tabular} \\ \hline
\multicolumn{1}{c|}{\textit{Neighborhood}} & \begin{tabular}[c]{@{}l@{}}N1 - Fraction of borderline points\\ T1 - Fraction of hyper-spheres covering cardinality\\ LSC - Local set average cardinality\end{tabular} \\ \hline
\multicolumn{1}{c|}{\textit{Network}} & \begin{tabular}[c]{@{}l@{}}Density \\ Hubs\end{tabular} \\ \hline
\multicolumn{1}{c|}{\textit{Dimensionality}} & \begin{tabular}[c]{@{}l@{}}T2 - Average number of features per dimension\\ T3 - Average number of PCA dimensions\\ T4 - Ratio of PCA dimension to original dimension\end{tabular} \\ \hline
\multicolumn{1}{c|}{\textit{Class Imbalance}} & \begin{tabular}[c]{@{}l@{}}C1 - Entropy of classes proportions\\ C2 - Imbalance ratio\end{tabular} \\ 
\toprule
\\
\multicolumn{2}{c}{\textbf{Landmarking Features}} \\ 
\\
\toprule

\multicolumn{1}{c|}{\textit{Linear }} & \begin{tabular}[c]{@{}l@{}}\textit{linear\_discr} - Score of Linear Discriminant classifier \end{tabular} \\ \hline

\multicolumn{1}{c|}{\textit{Naive bayes}} & \begin{tabular}[c]{@{}l@{}}\textit{naive\_bayes} - Score of Naive Bayes classifier \end{tabular} \\ \hline
\multicolumn{1}{c|}{\textit{Nearest Neighbour}} & \begin{tabular}[c]{@{}l@{}}\textit{elite\_nn} - Score of Elite Nearest Neighbor \\ \textit{one\_nn} - Score of the 1-Nearest Neighbor classifier \end{tabular} \\ \hline

\multicolumn{1}{c|}{\textit{Decision Tree}} & \begin{tabular}[c]{@{}l@{}}\textit{best\_node} - Score of the best single decision tree node\\ 
\textit{random\_node} - Score of a single node model induced by a random attribute. \\\textit{worst\_node}  - Score of a single node model induced by the worst attribute\end{tabular} \\ 
\bottomrule
\end{tabular}
\vspace{5pt}
\caption{Statistical \& Information-theoretic and Landmarking meta-features}
\label{tab:stats-info-features}
\end{table}
        
There is no straightforward method to compare two data characteristics unless we use a downstream task to evaluate extrinsically. We describe our experiment below in steps, which compares the effectiveness of clustering indices against landmarking, statistical and information-theoretic features through the performance evaluation of an extrinsic regression task.
\begin{itemize}
    \item Pick a subset of datasets from our experiment and frame a 5-fold cross-validation on the datasets
    \item Every dataset in the fold shall undergo subsampling individually
    \item Prepare the model-fitness scores (dependent variable) for each subsample
    \item Clustering Indices
    \begin{itemize}
        \item Prepare the clustering indices (features) from Table \ref{tab:indices_list} for each subsample 
        \item Learn regressors for each model class from clustering indices to model-fitness
    \end{itemize}
    \item Classic Statistical Features
    \begin{itemize}
        \item Collect a list of statistical \& information-theoretic features \cite{article,komorniczak2022complexity} as listed in Table~\ref{tab:stats-info-features}
        \item Generate the statistical features for each subsample
        \item Learn regressors for each model class from statistical features to model-fitness
    \end{itemize}
    \item Landmarking Features
    \begin{itemize}
        \item Collect a list of landmarking features \cite{PYMFEJMLR:v21:19-348,pymfe} as listed in Table~\ref{tab:stats-info-features}
        \item Generate the landmarking features for each subsample
        \item Learn regressors for each model class from landmarking features to model-fitness
    \end{itemize}    
    \item Measure the average cross-validation performance using $R^2$ score for the testing and training folds across model classes
\end{itemize}

\begin{table}
\centering
\begin{tabular}{c|cc|cc|cc}
\toprule
& \multicolumn{2}{c|}{\textbf{Clustering Indices}} & \multicolumn{2}{c|}{\textbf{Statistical Features}} & \multicolumn{2}{c}{\textbf{Landmarking}} \\
\textbf{Methods} & {Training} & {Testing} & {Training} & {Testing} & {Training} & {Testing}\\
\midrule
\textit{DT} & 0.90 & 0.89 & 0.75 & 0.57 & 0.61 & 0.48\\
\textit{RF} & 0.91 & 0.91 & 0.76 & 0.58 & 0.62 & 0.48\\
\textit{LR} & 0.89 & 0.89 & 0.77 & 0.61 & 0.62 & 0.59\\
\textit{KNN} & 0.90 & 0.89 & 0.75 & 0.59 & 0.61 & 0.44\\
\textit{XG} & 0.89 & 0.89 & 0.72 & 0.53 & 0.59 & 0.49\\
\textit{SVC} & 0.88 & 0.88 & 0.74 & 0.58 & 0.60 & 0.33\\
\hline
\textbf{Mean} & \textbf{0.90} & \textbf{0.89} & \textbf{0.75} & \textbf{0.58} & \textbf{0.63} & \textbf{0.47}\\
\bottomrule
\end{tabular}
\vspace{5pt}
\caption{Regressor $R^2$ performance comparison. The columns list the cross-validated training and testing performance of regressors when we use cluster indices, statistical \& information-theoretic, and landmarking meta-features from the literature, respectively.}
\label{tab:clus_vs_prob}
\end{table}

Table \ref{tab:clus_vs_prob} summarizes the performance of the extrinsic regression task through i) clustering indices features, ii) classic statistical features, and iii) landmarking. It is apparent from the table that the clustering indices features are better than the other two strategies for capturing the dataset characteristics, at least with respect to an extrinsic regression task.  The performance numbers show that the generalization error is higher for classic statistical features, although the training performance is reasonable. Landmarking, on the other hand, is unable to capture the dataset characteristics during training. As a result, the landmarking features generalize poorly during validation.

\begin{table}
\centering
\begin{tabular}{cccc|cccccc} 
\toprule
\multicolumn{4}{c|}{\textbf{Choice of Clustering Indices}} & \multicolumn{1}{l}{\textbf{DT}} & \multicolumn{1}{l}{\textbf{RF}} & \multicolumn{1}{l}{\textbf{LR}} & \multicolumn{1}{l}{\textbf{KNN~}} & \multicolumn{1}{l}{\textbf{SVC}} & \multicolumn{1}{l}{\textbf{XG}} \\ 
\midrule
 \multicolumn{2}{c}{Internal Indices}& \multicolumn{2}{c|}{{\cellcolor[rgb]{0.853,0.853,0.853}}\sout{External Indices}} & 0.58 & 0.47 & 0.35 & 0.38 & 0.51 & 0.76\\
 \hline
  \multicolumn{2}{c}{{\cellcolor[rgb]{0.853,0.853,0.853}}\sout{Internal Indices}}& \multicolumn{2}{c|}{External Indices} & 0.77 & 0.82 & 0.81 & 0.73 & 0.74 & 0.77 \\
\hline
\midrule

K-means~ & Hierarchical & Spectral & {\cellcolor[rgb]{0.853,0.853,0.853}}\sout{HDBSCAN} & 0.72 & 0.84 & 0.70 & 0.72 & 0.67 & 0.78  \\
K-means~ & Hierarchical & {\cellcolor[rgb]{0.853,0.853,0.853}}\sout{Spectral} & HDBSCAN & 0.81 & 0.85 & 0.75 & 0.75 & 0.76 & 0.79  \\
K-means~ & {\cellcolor[rgb]{0.853,0.853,0.853}}\sout{Hierarchical} & Spectral & HDBSCAN & 0.81 & 0.86 & 0.74 & 0.76 & 0.76 & 0.81  \\
{\cellcolor[rgb]{0.853,0.853,0.853}}\sout{Kmeans~} & Hierarchical & Spectral & HDBSCAN & 0.79 & 0.83 & 0.74 & 0.73 & 0.73 & 0.79  \\ 
\hline
K-means~ & Hierarchical & {\cellcolor[rgb]{0.853,0.853,0.853}}\sout{Spectral} & {\cellcolor[rgb]{0.853,0.853,0.853}}\sout{HDBSCAN} & 0.79 & 0.81 & 0.66 & 0.74 & 0.69 & 0.74  \\ 
K-means~ & {\cellcolor[rgb]{0.853,0.853,0.853}}\sout{Hierarchical} & {\cellcolor[rgb]{0.853,0.853,0.853}}\sout{Spectral} & HDBSCAN & 0.8 & 0.86 & 0.76 & 0.74 & 0.73 & 0.80 \\
{\cellcolor[rgb]{0.853,0.853,0.853}}\sout{K-means~} & {\cellcolor[rgb]{0.853,0.853,0.853}}\sout{Hierarchical} & Spectral & HDBSCAN & 0.78 & 0.80 & 0.73 & 0.73 & 0.75 & 0.77 \\ 
\hline
K-means~ & {\cellcolor[rgb]{0.853,0.853,0.853}}\sout{Hierarchical} & {\cellcolor[rgb]{0.853,0.853,0.853}}\sout{Spectral} & {\cellcolor[rgb]{0.853,0.853,0.853}}\sout{HDBSCAN} & 0.76 & 0.79 & 0.62 & 0.74 & 0.64 & 0.76  \\
{\cellcolor[rgb]{0.853,0.853,0.853}}\sout{K-means~} & Hierarchical & {\cellcolor[rgb]{0.853,0.853,0.853}}\sout{Spectral} & {\cellcolor[rgb]{0.853,0.853,0.853}}\sout{HDBSCAN} & 0.76 & 0.80 & 0.62 & 0.73 & 0.65 & 0.62 \\
{\cellcolor[rgb]{0.853,0.853,0.853}}\sout{K-means~} & {\cellcolor[rgb]{0.853,0.853,0.853}}\sout{Hierarchical} & Spectral & {\cellcolor[rgb]{0.853,0.853,0.853}}\sout{HDBSCAN} & 0.69 & 0.68 & 0.72 & 0.67 & 0.61 & 0.67 \\
{\cellcolor[rgb]{0.853,0.853,0.853}}\sout{K-means~} & {\cellcolor[rgb]{0.853,0.853,0.853}}\sout{Hierarchical} & {\cellcolor[rgb]{0.853,0.853,0.853}}\sout{Spectral} & HDBSCAN & 0.77 & 0.80 & 0.74 & 0.71 & 0.72 & 0.66  \\
\midrule
K-means~ & Hierarchical & Spectral & HDBSCAN & \textbf{0.82} & \textbf{0.87} & \textbf{0.85} & \textbf{0.80} & \textbf{0.81} & \textbf{0.83} \\

  \bottomrule
\end{tabular}
\vspace{5pt}
\caption{Performance of the \emph{Mapper} module for different configurations of clustering indices.  \colorbox[rgb]{0.853,0.853,0.853}{\sout{Method}} indicates the ablation of the clustering indices from that specific clustering method while building the Mapper module. The \emph{Mapper} model performs best when we use all clustering indices (\emph{internal} \& \emph{external}) from all four families of clustering methods.}
\label{tab:ablation}
\end{table}

\subsection{Ablation Study}
Clustering indices reflect the ability of a clustering algorithm to form meaningful clusters over a dataset. A clustering algorithm may inherently suffer shortcomings due to its hypothesis and algorithmic limitations. For example, K-means cannot form nonconvex clusters, but methods such as Spectral clustering overcome this shortcoming. Likewise, Hierarchical clustering is very sensitive to outliers, but density-based methods such as HDBSCAN are resilient against outliers. This mentioned observation sets the premise for extracting clustering indices of a dataset using different clustering methods and concatenating them to create extended feature representation. 

We perform our ablation study by dropping clustering indices from one or more specific clustering methods.  We ablate the clustering indices to assess their relative contribution to an optimal build of the \emph{Mapper} module. We also extend our ablation study to measure the contribution of internal vs. external cluster indices listed in Table \ref{tab:indices_list}. The performance of the Mapper module for different configurations of cluster indices is summarized in Table \ref{tab:ablation}, where we observe that the \emph{Mapper} model performs best when we use all the \emph{internal} \& \emph{external} clustering indices from all four families of clustering methods.

\begin{table}
\centering
\scalebox{1}{%
\begin{tabular}{c|c}
\toprule
{\textbf{ Clustering index}} & \textbf{\ \ Spearman Coefficient\ \ } \\ 
\midrule
 {C-index - Kmeans \cite{c-index}}&   {-0.40} \\ 
 {Russel Rao - Spectral \cite{russel-rao}}&  {-0.51} \\ 
 {Purity - Spectral \cite{purity}}&  {-0.49}   \\ 
 {Entropy - Kmeans }&  {0.35}   \\ 
 {Dunn - Hierarchical \cite{dunn}}&  {0.31}   \\ 
 {Silhouette - Spectral \cite{sillhoutte}} &  {-0.46}    \\ 
 {Davies Bouldin - Spectral \cite{davies-bouldin}} &  {0.47} \\ 
 {Precision - HDBSCAN} &  {-0.43} \\ 
 {Completeness - HDBSCAN \cite{completeness}}&  {0.51}\\ 
 {PHI - Spectral \cite{clustercrit}}&  {0.28}   \\ 
\bottomrule 
\end{tabular}%
}
\vspace{5pt}
\caption{Top 10 critical features estimated using \emph{Spearman} correlation score estimation between the clustering indices and the estimated classification performance on a dataset. The top features clearly indicate that they are spread across all the variations of clustering indices as analyzed by the Ablation study.}
\label{tab:hardness-corr}
\end{table}
We further verify the result of the ablation study by observing the correlation between the feature importance and model-fitness. We extract the feature importance in terms of \emph{gain} score \cite{gainscore} from the tree-based regressor (XGBoost) that learns the mapping between the clustering indices feature and the model-fitness. We analyze the correlation between the extracted high-gain cluster indices and the expected classification performance to estimate the magnitude and direction of the influence.  Table \ref{tab:hardness-corr} narrates the list of critical clustering indices (features) that influence the dataset model-fitness estimated using \emph{Spearman} correlation measure. It is reassuring from the table that the critical features are distributed across all the variations of clustering indices, as suggested by the ablation study.

\subsection{Evaluating End-to-End \emph{CIAMS} System} \label{model-selection-validation}

We successfully validate our hypothesis that classification performance is a function of the dataset clustering indices for every model class by evaluating the performance of the \emph{Mapper} module of \emph{CIAMS}.  We now study the end-to-end \emph{CIAMS} system for the correctness of model-class recommendations given a test dataset. We use a $20-80$ train-test split to mimic the real-world production scenario where we get a limited labeled dataset to build models. This splitting also gives us an insight into the ability of the model to generalize with the availability of limited labeled data. As explained in Section \ref{evalphase}, we validate the \emph{CIAMS} system in both \emph{Single-shot} and \emph{Subsamples} mode. Given a test dataset $D'$, we generate the feature vector $\mathbf{I}'$ in the clustering indices feature space $\mathcal{I}$ using the procedure in Section \ref{data_construction} as $\mathbb{F}(D';\mathcal{A}):D'\rightarrow\mathbf{I}'$. We supply the clustering indices feature vector  $\mathbf{I}'$ as input to regressors $R_j \in \mathcal{R}$ to predict the classification performance ${F_1}_j^{pred}$ as $\mathbf{O}'^{(j)}\leftarrow R_j(\mathbf{I}';C_j)$ for every model class $C_j$. In the subsampling mode, the resulting $F_1^{pred}$ scores are of dimension $b\times c$, where $b$ is the number of subsamples drawn from the dataset $D'$ and $c$ is the number of model classes. In the \textit{Single-shot} mode, the dimension is $1\times c$.  The \emph{Subsamples} model output is collapsed using Equation \ref{expf1} into a $1\times c$ vector. 

We now have the predicted model-fitness $F_1$ scores for the test dataset $D'$ for each model class $C_j\in\mathcal{C}$.  Sorting the model fitness $F_1$ scores in descending order gets us the ranked recommendation of suitable model classes for the given test dataset $D'$.    We call the \emph{top3} model classes as $C_A^{pred}, C_B^{pred}, C_C^{pred}$ in the higher to lower classification performance order ($C_A^{} \ge C_B^{} \ge C_C^{}$). We compare the predicted rank order $C_A^{pred}, C_B^{pred}, C_C^{pred}$ with the true rank order $C_A^{true}, C_B^{true}, C_C^{true}$.  To get the ``true'' rank order of the model classes, we build tuned classifier models for all the model classes using the same dataset $D'$ that we feed to the \emph{Recommendation} pipeline.  We tune all the classification models through cross-validation. We use the evaluation score for the model classes to rank-order the classifiers, which in turn yields us $C_A^{true}, C_B^{true}, C_C^{true}$.

\subsubsection{Validation}
We validate the performance of \emph{CIAMS} using the corpus of \emph{60} datasets. Table \ref{tab:rankorder} presents the confusion matrix of the predicted and actual \emph{top3} model class recommendation by the \emph{CIAMS} system in both the \emph{Single-shot} and \emph{Subsampling} modes. Although the ranks are not exact matches, we observe a strong overlap between the true and the predicted \emph{top3} model classes. From the table, we observe that \emph{CIAMS} can recall (highlighted in \colorbox{yellow}{YELLOW}) the true \emph{top1} model class among the predicted \emph{top3} model class with recall scores of 78\% and 74\% in the \emph{Single-shot} and \emph{Sub-sampling} modes respectively.  Likewise, the \emph{top1} prediction from \emph{CIAMS} can recall the true \emph{top3} model classes with recall scores of 74\% and 70\% in the \emph{Single-shot} and \emph{Sub-sampling} modes, respectively.  The ability to recall the \emph{top1} model class in three-fourths of the test datasets is remarkable.

\begin{table}[!ht]
\centering
\begin{tabular}{l|ccc|c|c|ccc|c|c}
\toprule
\multicolumn{1}{c|}{\textbf{\textit{}}} & \multicolumn{5}{|c|}{\textit{\textbf{Single-shot (60-fold)}}} & \multicolumn{5}{c}{\textit{\textbf{Subsamples (6-fold $\times$6)}}}\\
\vspace{4pt}
 & $\ C_A^{pred}$ & $\ C_B^{pred}$ & $\ C_C^{pred}\ $ & Total & Hit & $\ C_A^{pred}$ & $\ C_B^{pred}$ & $C_C^{pred}\ $ & Total & Hit\\ 
\midrule
$C_A^{true}$ & 19 & 16 & 12 & 47 & \colorbox{yellow}{78\%} & 157 & 68 & 41 & 266 & \colorbox{yellow}{74\%}\\
$C_B^{true}$ & 12 & 10 & 12 & 34 & 55\% & 66 & 73 & 82 & 221 & 61\%\\
$C_C^{true}$ & 13 & 9 & 8 & 30 & 50\% & 28 & 66 & 58 & 152 & 42\%\\
\midrule
Total & 44 & 35 & 32 & \emph{60} & &  251 & 207 & 181 & \emph{362} & \\
\midrule
Hit & \colorbox{yellow}{74\%} & 58\% & 53\% & & &  \colorbox{yellow}{70\%} & 58\% & 50\% & & \\

\bottomrule
\end{tabular}%
\vspace{5pt}
\caption{Confusion matrix of the true and predicted \emph{top3} model classes in \emph{Single-shot} and  \emph{Subsampling} modes.}
\label{tab:rankorder}
\end{table}

\begin{table}
\centering
\begin{tabular}{l|cc|ccc|c}
\toprule
\textbf{Dataset} &\textbf{ Dims} & \textbf{Count} & $C_A^{pred}$ & $C_B^{pred}$ & $C_C^{pred}$ & $C_{voting}^{pred}$ \\  \midrule
australian & 15 & 689 & 0.61 & {\ul \textbf{0.84}} & \emph{0.83} & \emph{0.83} \\
bankruptcy & 7 & 250 & 0.51 & {\ul \textbf{0.97}} & \emph{0.96} & 0.92 \\
christine & 1637 & 5417 & 0.65 & {\ul \textbf{0.71}} & \emph{0.70} & \emph{0.70} \\
cleve & 14 & 302 & 0.68 & 0.66 & {\ul \textbf{0.78}} & \emph{0.77} \\
haberman & 4 & 305 & 0.32 & 0.35 & \emph{0.43} & \colorbox{yellow}{\ul \textbf{0.52}} \\
jm1 & 22 &  10879& 0.42 & {\ul \textbf{0.51}} & \emph{0.50} & 0.46 \\
mfeat-morphological & 7 & 1999 & 0.70 & 0.82 & {\ul \textbf{0.99}} & \emph{0.98} \\
monk-2 & 7 & 431 & 0.80 & {\ul \textbf{0.98}} & {\ul \textbf{0.98}} & {\ul \textbf{0.98}} \\
nba\_logreg & 20 & 1328 & {\ul \textbf{0.66}} & \emph{0.65} & 0.60 & 0.63 \\
no2 & 8 & 499 & 0.54 & 0.55 & {\ul \textbf{0.56}} & {\ul \textbf{0.56}} \\
ozone & 73 & 2533 & 0.37 & \emph{0.42} & 0.31 & \colorbox{yellow}{\ul \textbf{0.43}} \\
philippine & 309 & 5831 & \emph{0.72} & 0.68 & 0.66 & \colorbox{yellow}{\ul \textbf{0.73}} \\
phisingwbsite & 31 & 11054 & 0.92 & 0.92 & {\ul \textbf{0.95}} & {\ul \textbf{0.95}} \\
piechart3 & 38 & 1076 & 0.34 & {\ul \textbf{0.46}} & 0.20 & \emph{0.37} \\
saheart & 10 & 461 & 0.58 & 0.58 & {\ul \textbf{0.60}} & {\ul \textbf{0.60}} \\
scene\_urban & 300 & 2406 & \emph{0.95} & 0.90 & {\ul \textbf{0.97}} & \emph{0.95} \\
segment & 20 & 2309 & 0.97 & {\ul \textbf{0.98}} & 0.97 & {\ul \textbf{0.98}} \\
speech & 401 & 3685 & 0.34 & {\ul \textbf{0.46}} & 0.20 & \emph{0.37} \\
UniversalBank & 13 & 4999 & 0.41 & {\ul \textbf{0.51}} & \emph{0.48} & \emph{0.48} \\
wine4 & 12 & 1598 & \emph{0.12} & 0.03 & {\ul \textbf{0.13}} & \emph{0.12} \\
amazon\_emp\_access* &9  & 32769  &0.05 & \emph{0.18} & {\ul\textbf{ 0.26}} & 0.11 \\
 clickpred\_small*& 12  & 39948  & 0.15 & {\ul\textbf{0.43}} & 0.41 & \emph{0.42}\\
namoa*  & 120 & 34465  &  {\ul\textbf{0.94}}  & 0.91 & 0.92 & \emph{0.93} \\
\textit{FOREX\_eurgbp*}& 11 & 43825 & \emph{0.49}  & 0.37 & {\ul \textbf{0.50}} &{\ul\textbf{0.50}} \\
run\_or\_walk\_info* & 8 & 88588  & \emph{0.98} & \emph{0.98} &   {\ul\textbf{0.99}} & \emph{0.98} \\
\bottomrule
\end{tabular}%
\vspace{5pt}
\caption{Performance comparison of the \emph{top3} classifiers recommended by \emph{CIAMS}.  The best overall score is \textbf{bold-faced} and {\ul underlined}. The second-best score is in \emph{italics}. The performance of the Weighted Voting Classifier is higher than the \emph{top3} classifiers in three \colorbox{yellow}{highlighted} datasets.}
\label{tab:VCLF_CIAMS_subsampling}
\end{table}

\subsubsection{Testing}
We test \emph{CIAMS}'s \emph{top3} recommend model classes using a separate \emph{hold-out} set of public domain binary class datasets listed in the first column of Table \ref{tab:VCLF_CIAMS_subsampling}. We compare the actual classification performance of the \emph{top3} model classes and tabulate the results in Table \ref{tab:VCLF_CIAMS_subsampling}. We report the 5-fold cross-validation performance of the \emph{top3} classifiers on the test datasets.  The \emph{top3} recommended classifiers win in 22 of 25 datasets.  By running cross-validation, we use the best of \emph{top3} recommended classifiers for a given test dataset.  Essentially, we bring the complexity down from running an exhaustive model search to choosing the best from only three choices.  We use the best classifier from the recommended list as the underlying model of the end-to-end Automated ML system to expose a SaaS API for automatic dataset classification.  

As an exercise, we also ensemble the \emph{top3} classifiers into a Weighted Voting Classifier ${C}_{voting}$. We list the ensemble's performance in the last column of Table~\ref{tab:VCLF_CIAMS_subsampling}. The voting classifier works best for 9 of 25 datasets. When we consider the \emph{top2} scores from the column, we find that the voting classifier performs well for 21~of~25 datasets. The voting classifier abstracts the \emph{top3} models, simplifying the Automated ML pipeline with one final prediction model. Of the 9 (nine) best scores, we also observe that the voting classifier ensemble is marginally better than the constituent classifiers in 3 (three) datasets highlighted in YELLOW in Table \ref{tab:VCLF_CIAMS_subsampling}. As the performance improvement is not significant, we choose to skip ensembling the \emph{top3} model classes. 

\subsection{Validating CIAMS based end-to-end Automated ML System}
\label{validating_saas}

Automated Machine Learning platforms provide significant cost savings to businesses, focusing on complex processes such as product innovation, market penetration, and enhanced client satisfaction. Automated ML platforms decrease the energy spent on time and resource-consuming processes such as model selection, feature engineering, and hyper-parameter tuning. The major cloud players such as Microsoft and Amazon have their version of the Automated ML platforms. In responding to the demand for accessible and affordable automatic machine learning platforms, open-source frameworks are also available to put the data to use as quickly and with as little effort as possible. Table \ref{tab:automlplatforms} lists a limited set of commercial and open-source Automated ML platforms, with which we make a performance comparison against \emph{CIAMS} based Automated ML system.
\begin{table}
    \centering
    \begin{tabular}{c|c}
    \toprule
    \textbf{Automated ML Platform} & \textbf{Name}\\
    \midrule
    {Auto-sklearn } & \textbf{A-SKL}\\
    {Auto-WEKA } & \textbf{A-Weka}\\
    {TPOT } & \textbf{TPOT}\\
    {Microsoft Azure Automated ML } & \textbf{Azure}\\
    {Amazon SageMaker } & \textbf{SageMaker}\\
    {H2O AutoML } & \textbf{H2O}\\
    {FLAML } & \textbf{FLAML}\\
    \bottomrule
    \end{tabular}
    \vspace{5pt}
    \caption{A limited list of commercial and open-source Automated ML platforms.}
    \label{tab:automlplatforms}
\end{table}

Automated machine learning frameworks generally apply standardized techniques for feature selection, feature transformation, and data imputation on datasets developed over the years. However, the underlying methods used to automate machine learning tasks are different. We experimentally assess these methods in an end-to-end style across various datasets. We perform a quantitative comparison of the performance of \emph{CIAMS} measured using $F_1$ score with the other Automated ML candidate methods listed in Table \ref{tab:automlplatforms}.

Commercial and noncommercial Automated ML platforms apply different model ensembling techniques to boost performance. In our design, we use the \emph{top3} \emph{CIAMS} recommended model classes and build the classifiers using the \emph{labeled} part of the \emph{Production} dataset and tune the constituent classifiers using $5$-fold cross-validation. We compare the evaluation (or \emph{test}) set performance of the best-performing model among the \emph{top3} \emph{CIAMS} against the performance of the other Automated ML methods in Table \ref{tab:mean_f1_single-shot}.

\begin{table}[t]
    \centering
    \resizebox{\textwidth}{!}{
    \begin{tabular}{l|cccccccc}
\toprule \\
\textbf{Dataset} & \multicolumn{1}{l}{\textbf{CIAMS}} & \multicolumn{1}{l}{\textbf{FLAML}} & \multicolumn{1}{l}{\textbf{H2O}} & \multicolumn{1}{l}{\textbf{TPOT}} & \multicolumn{1}{l}{\textbf{A-WEKA}} & \multicolumn{1}{l}{\textbf{AzureML}} & \multicolumn{1}{l}{\textbf{SageMaker}} & \multicolumn{1}{l}{\textbf{A-SKL}} \\ \midrule
\textit{australian} & \textit{0.84} & \textit{0.84} & \textbf{0.85} & \textit{0.84} & 0.82 & {\ul \textbf{0.87}} & - & \textbf{0.85} \\
\textit{bankruptcy} & \textbf{0.97} & {\ul \textbf{0.98}} & \textbf{0.97} & \textit{0.94} & 0.91 & - & - & 0.75 \\
\textit{christine} &{\ul \textbf{0.71} }& \textit{0.68} & {\ul \textbf{0.71}} & \textbf{0.70} & 0.65 & \textbf{0.70} & \textit{0.68}& \textbf{0.70} \\
\textit{cleve} &{\ul \textbf{0.78}} & 0.70 & \textbf{0.77} & {\ul \textbf{0.78}} & 0.68 & 0.27 & - & \textit{0.75} \\
\textit{haberman} & \ \textit{0.43} & 0.38 & 0.40 & {\ul \textbf{0.55}} & \textbf{0.47} & 0.26 & - & 0.32 \\
\textit{jm1} & \textbf{0.51} & 0.36 & \textit{0.50} & 0.45 & 0.43 & 0.37 & {\ul \textbf{0.55}} & 0.22 \\
\textit{mfeat-morphological} & {\ul \textbf{0.99}} & {\ul \textbf{0.99}} & {\ul \textbf{0.99}} & {\ul \textbf{0.99}} & {\ul \textbf{0.99}} & {\ul \textbf{0.99}} & - & {\ul \textbf{0.99}} \\
\textit{monk-2} & \textit{0.98} & \textbf{0.99} & \textit{0.98} & \textit{0.98} & {\ul \textbf{1.00}} & \textbf{0.99} & - & {\ul \textbf{1.00}} \\
\textit{nba\_logreg} & \textbf{0.66} & 0.57 & 0.62 & \textit{0.65} & {\ul \textbf{0.67}} & {\ul \textbf{0.67}} & - & \textit{0.65} \\
\textit{no2} & \textbf{0.56} & {\ul \textbf{0.58}} & 0.50 & \textbf{0.56} & \textit{0.55} & {\ul \textbf{0.58}} & - & {\ul \textbf{0.58}} \\
\textit{ozone} &  \textbf{0.42} & 0.29 & \textbf{0.42} & 0.37 & 0.37 & 0.20 & {\ul \textbf{0.43}} & \textit{0.40} \\
\textit{philippine} &  \textbf{0.72} & {\ul \textbf{0.73}} & 0.68 & {\ul \textbf{0.73}} & 0.58 & {\ul \textbf{0.73}} & \textit{0.71} & \textbf{0.72} \\
\textit{phisingwbsite} & \textbf{0.95} & \textbf{0.95} & {\ul \textbf{0.96}} & \textit{0.94} & \textit{0.94} & {\ul \textbf{0.96}} & 0.93 & \textbf{0.95} \\
\textit{piechart3} & {\ul \textbf{0.46}} & 0.31 & 0.27 & \textbf{0.39} & 0.30 & 0.23 & - & \textit{0.36} \\
\textit{saheart} & \textbf{0.60} & \textbf{0.60} & \textbf{0.60} & 0.57 & 0.56 & \textit{0.59} & - & {\ul \textbf{0.62}} \\
\textit{scene\_urban} & {\ul \textbf{0.97}} & \textbf{0.92} & 0.56 & {\ul \textbf{0.97}} & 0.33 & \textit{0.91} & - & {\ul \textbf{0.97}} \\
\textit{segment} & \textbf{0.98} & \textbf{0.98} & \textbf{0.98} & \textbf{0.98} & \textit{0.96} & {\ul \textbf{0.99}} & - & {\ul \textbf{0.99}} \\
\textit{speech} & {\ul \textbf{0.46}} & 0.02 & 0.27 & \textbf{0.39} & 0.04 & 0.02 & 0.15 & \textit{0.36} \\
\textit{UniversalBank} & {\ul \textbf{0.51}} & \textit{0.47} & \textbf{0.49} & \textbf{0.49} & 0.43 & 0.45 & 0.36 & 0.45 \\
\textit{wine4} & {\ul \textbf{0.13}} & \textbf{0.06} &\textit{ 0.03} & \textit{0.03} & \textbf{0.06} &\textit{ 0.03} & - & {\ul \textbf{0.13}} \\
\textit{amazon\_emp\_access*} &\textbf{ 0.26} & 0.05 & \textit{0.23} & 0.22 & 0.15 & 0.15 & {\ul\textbf{0.37}} &  \textbf{0.26}\\
\textit{clickpred\_small*}    & \textbf{0.43} & 0.15 & {\ul\textbf{0.44}} & \textit{0.37} & 0.21 & 0.16 & - &  0.22\\
\textit{namao*}               & {\ul\textbf{0.94}} & 0.21 & {\ul\textbf{0.94}} & \textbf{0.93} & \textit{0.92} & {\ul\textbf{0.94}} & 0.91 &  \textbf{0.93}\\
\textit{FOREX\_eurgbp*}       & \textbf{0.50} & 0.34 & \textit{0.45} & \textbf{0.50} & \textbf{0.50} & \textbf{0.50}& - & {\ul\textbf{0.51}}\\
\textit{run\_or\_walk\_info*} & {\ul\textbf{0.99}} & 0.34 & \textbf{0.98} & \textbf{0.98} & \textit{0.95} & \textbf{0.98} & - &  \textbf{0.98}\\

\midrule
\textbf{\#Top1 Performance} & {\ul\textbf{10}} & 4 & \emph{5} & \emph{5} & 3 & \textbf{8} & 3 & \textbf{8} \\
\textbf{\#Top2 Performance} & {\ul\textbf{22}} & 10 & {12} & \emph{14} & 5 & {5} & 3 & \textbf{15} \\
\textbf{\#Top3 Performance} & {\ul\textbf{25}} & 12 & {17} & \textbf{21} & 11 & {15} & 4 & \emph{20} \\
\textbf{Average Rank} & {\ul\textbf{1.68}} & 3.36 & 2.64 & \textbf{2.4} & 3.64 & 3.3 & 3.11 & \emph{2.56} \\
\bottomrule
\end{tabular}
}
\vspace{5pt}
\caption{Weighted $F_1$ score achieved by different Automated ML methods for various public domain binary class datasets (as \emph{production} datasets) in the Single-shot setting. The best score is presented as {\ul \textbf{underlined \& bold-faced}}, the second best is \textbf{bold-faced} and the next best is \textit{italicized}. The average rank is the average of the rank order position scored by each Automated ML method, where a low number means higher-performance. Datasets with a * marker are larger in size.}
\label{tab:mean_f1_single-shot}
\end{table}

While comparing the performance of \emph{CIAMS} against other Automated ML methods, we provide the test dataset in full as input to all the systems evaluated in this experiment. We observe from Table \ref{tab:mean_f1_single-shot} that \emph{CIAMS} is winning against the other methods with an average rank of \emph{1.68}, followed by Auto-weka scoring an average of \emph{2.4}.  It is interesting to observe the \emph{CIAMS} method scoring a definite \emph{top3} position in all \emph{25} test datasets. The clear win also reflects in the number of \emph{top2} positions at \emph{22} of 25. This observation gives us a strong validation that \emph{CIAMS} based end-to-end Automated ML system is at par if not better than other commercial and open-source automated machine learning methods even without any explicit feature engineering incorporated by the other methods.


\begin{table}
\centering
\begin{tabular}{l|ccccc|c}
\toprule
\textbf{Dataset} & \textbf{FLAML} & \textbf{H2O} & \textbf{TPOT} & \textbf{A-WEKA} & \textbf{A-SKL} &  \\ \midrule
\textit{australian}           & WIN                  & WIN  & LOSE & WIN   & WIN  &        \\
\textit{bankruptcy}           & WIN                  & WIN  & WIN  & WIN   & WIN  &        \\
\textit{christine}            & WIN                  & WIN  & WIN  & WIN   & LOSE &        \\
\textit{cleve}                & LOSE                 & LOSE & LOSE & WIN   & LOSE &        \\
\textit{haberman}             & WIN                  & WIN  & LOSE & WIN   & WIN  &        \\
\textit{jm1}                  & WIN                  & LOSE & WIN  & WIN   & WIN  &        \\
\textit{mfeat-morphological}  & WIN                  & LOSE & LOSE & WIN   & WIN  &        \\
\textit{monk-2}               & WIN                  & WIN  & WIN  & WIN   & LOSE &        \\
\textit{nba\_logreg}          & WIN                  & WIN  & WIN  & WIN   & WIN  &        \\
\textit{no2}                  & WIN                  & WIN  & WIN  & WIN   & WIN  &        \\
\textit{ozone}                & WIN                  & LOSE & LOSE & WIN   & WIN  &        \\
\textit{philippine}           & WIN                  & WIN  & WIN  & WIN   & WIN  &        \\
\textit{phisingwbsite}        & WIN                  & WIN  & WIN  & WIN   & LOSE &        \\
\textit{piechart3}            & WIN                  & WIN  & WIN  & WIN   & WIN  &        \\
\textit{saheart}              & WIN                  & LOSE & LOSE & WIN   & LOSE &        \\
\textit{scene\_urban}         & WIN                  & WIN  & WIN  & WIN   & WIN  &        \\
\textit{segment}              & WIN                  & WIN  & WIN  & WIN   & WIN  &        \\
\textit{speech}               & WIN                  & LOSE & WIN  & WIN   & WIN  &        \\
\textit{UniversalBank}        & WIN                  & WIN  & WIN  & -     & WIN  &        \\
\textit{wine4}                & WIN                  & LOSE & LOSE & WIN   & WIN  &        \\
\textit{amazon\_emp\_access*} & WIN                  & WIN  & WIN  & WIN   & WIN  &        \\
\textit{clickpred\_small*}    & WIN                  & LOSE & LOSE & WIN   & WIN  &        \\
\textit{namao*}               & LOSE                 & LOSE & WIN  & WIN   & LOSE &        \\
\textit{FOREX\_eurgbp*}       & WIN                  & WIN  & LOSE & WIN   & LOSE &        \\
\textit{run\_or\_walk\_info*} & WIN                  & WIN  & WIN  & -     & WIN  &        \\ \midrule
\textit{}                     &                      &      &      &       &      &\textbf{ AVG }   \\\midrule
\textbf{WIN \%}               & 92\%                 & 64\% & 64\% & 100\% & 72\% & 78.4\% \\
\textbf{LOSE \%}              & 8\%                  & 36\% & 36\% & 0\%   & 28\% & 21.6\% \\ \bottomrule
\end{tabular}

\vspace{5pt}
\caption{Student's $t$-test results with a significance level of $0.02$ for the comparison of \emph{CIAMS} against other Automated ML methods.}.
\label{tab:t_test_res}
\end{table}

We further study the statistical significance of the performance of \emph{CIAMS} against other Automated ML methods using two samples $t$-test in Table \ref{tab:t_test_res}. It is evident from the table that \emph{CIAMS} wins over the other methods in an average of three-fourths (78\%) of the test datasets. The next best methods in the comparison are TPOT and H2O, where \emph{CIAMS} wins in two-thirds (64\%) of the test datasets population. It is remarkable to observe \emph{CIAMS} ruling over FLAML, and Auto-WEKA with 92\% and 100\% wins. From Tables \ref{tab:mean_f1_single-shot} and \ref{tab:t_test_res}, we conclude that \emph{CIAMS} is a great contender for becoming an Automated ML system for dataset classification in production settings.

\begin{table} 
\centering
\resizebox{\textwidth}{!}{%
\begin{tabular}{l|cccccccc}
\toprule \\
\textbf{Dataset} & \multicolumn{1}{l}{\textbf{CIAMS}} & \multicolumn{1}{l}{\textbf{FLAML}} & \multicolumn{1}{l}{\textbf{H20}} & \multicolumn{1}{l}{\textbf{TPOT}} & \multicolumn{1}{l}{\textbf{A-WEKA}} & \multicolumn{1}{l}{\textbf{AzureML}} & \multicolumn{1}{l}{\textbf{SageMaker}} & \multicolumn{1}{l}{\textbf{A-SKL}} \\ \midrule
\textit{australian} & {\ul \textbf{10.9}} & \textbf{11.1} & 13.8 & 30.5 & 184.0 & 1064.6 & - & 3745.3 \\
\textit{bankruptcy} & {\ul \textbf{9.2}} & \textbf{9.4} & 34.3 & 30.0 & 197.0 & - & - & 4699.3 \\
\textit{christine} & 586.8 & 606.9 & 1181.7 & {\ul \textbf{209.6}} & \textbf{290.0 }& 1273.4 & 2832.7 & 3705.9 \\
\textit{cleve} & {\ul \textbf{9.7}} & \textbf{9.9} & 101.9 & 65.8 & 196.0 & 1111.5 & - & 4699.3 \\
\textit{haberman} & {\ul \textbf{16.3}} & \textbf{16.6} & 18.2 & 28.8 & 206.0 & 1089.3 & - & 3668.3 \\
\textit{jm1} & 388.8 & 3120.5 & {\ul \textbf{35.2}}& \textbf{157.2} & 178.0 & 1110.2 & 1835.7 & 3605.0 \\
\textit{mfeat-morphological} & {\ul \textbf{20.9}} & \textbf{21.1} & 34.5 & 33.2 & 171.0 & 1102.0 & - & 3644.9 \\
\textit{monk-2} & {\ul \textbf{9.3}} & \textbf{9.4} & 24.6 & 24.0 & 169.0 & 1128.1 & - & 3811.9 \\
\textit{nba\_logreg} & {\ul \textbf{19.1}} & \textbf{19.2} & 14.2 & 29.1 & 165.0 & 1156.2 & - & 3664.9 \\
\textit{no2} & {\ul \textbf{9.8}} & \textbf{10.0} & 19.0 & 35.0 & 173.0 & 1117.0 & - & 3641.8 \\
\textit{ozone} & 71.5 & 71.9 &  {\ul\textbf{51.5}} & \textbf{70.3} &  212.0 & 1152.9 & 1805.6 & 3620.4 \\
\textit{philippine} & 534.4 & 536.1 & {\ul\textbf{156.4}} & 358.9 &\textbf{ 202.0} & 1179.3 & 1986.9 & 3604.5 \\
\textit{phisingwbsite} & 367.0 & 3224.0 & {\ul\textbf{45.3}} & \textbf{67.9}& 417.0 & 1131.8 & 1745.4 & 3621.4 \\
\textit{piechart3} & \textbf{24.4} & 24.7 & {\ul \textbf{20.1}} & 25.3 & 180.0 & 1141.4 & - & 3706.5 \\
\textit{saheart} & {\ul \textbf{15.6}} & \textbf{15.7} & 16.8 & 30.7 & 184.0 & 1133.5 & - & 3716.7 \\
\textit{scene\_urban} & {\ul \textbf{93.4}} & \textbf{94.0} & 111.4 & 210.2 & 180.0 & 1165.7 & - & 3618.7 \\
\textit{segment} & {\ul \textbf{32.0}} & \textbf{32.3} & 68.4 & 29.5 & 158.0 & 1092.4 & - & 3741.8 \\
\textit{speech} & \textbf{24.4} & 25.3 & {\ul \textbf{20.1}} & 25.3 & 197.0 & 1170.7 & 2168.1 & 3706.5 \\
\textit{UniversalBank} & 268.6 & 268.7 & {\ul \textbf{19.2}} &  \textbf{64.9} & 170.0 & 1154.1 & 1835.7 & 3611.9 \\
\textit{wine4} & {\ul \textbf{18.0}} & \textbf{18.2} & 19.3 & 26.3 & 188.0 & 1161.3 & - & 7188.5 \\ 
\textit{amazon\_emp\_access*} & 340.7 & 341.1 & \textbf{53.0 }&  {\ul\textbf{86.7}} & 171.9 & 1150.9 & 2046.8 & 3624.7 \\
\textit{clickpred\_small*}    & 278.3 & 279.2 & {\ul\textbf{53.7}} & \textbf{63.7} &5139.0 & 1131.4 & - & 3614.6 \\
\textit{namao*}               & 649.7 & 651.4 & 528.7 & {\ul\textbf{124.9}} & \textbf{325.0} & 1137.7 & 2077.5 & 3737.3 \\
\textit{FOREX\_eurgbp*}       & 403.7 & 515.7 & {\ul\textbf{50.8}} & \textbf{117.7}& 205.0 & 1112.8 & - & 3754.0 \\
\textit{run\_or\_walk\_info*} & 615.2 & 403.9 & \textbf{188.0} & {\ul\textbf{83.4}} & 13920.0 & 1145.4 & - & 3648.7 \\
\midrule
\#Fastest Performance & {\ul\textbf{12}} & 0 & \textbf{9} & \emph{4} & 0 & 0 & 0 & 0 \\
\#Faster Performance & {\ul\textbf{14}} & \textbf{12} & \textit{11} & {10} & 3 & 0 & 0 & 0 \\
\bottomrule
\end{tabular}
}
\vspace{5pt}
\caption{Comparison of time taken (in seconds) by \emph{CIAMS} and other Automated ML methods for different binary class datasets. The shortest time is presented as {\ul \textbf{underlined \& bold-faced}} and the second shortest is \textbf{bold-faced}. Datasets with a *~marker are larger in size.}
\label{tab:CIAMS-timing}
\end{table}


Table \ref{tab:CIAMS-timing} shows the time taken by each of the automated machine learning methods for building models and making predictions, end-to-end.  Amazon SageMaker and Auto-sklearn are the slowest methods consuming over an hour for each dataset.  Azure Automated ML is the next slowest method, consuming over 15 mins for each dataset on average. TPOT is reasonably faster, with an average time of less than a minute.  FLAML and H2O AutoML are the second fastest methods to make \emph{CIAMS} the fastest method for automated machine learning in a limited dataset experiment. \emph{CIAMS} scores the \emph{top1} fastest position on \emph{12} and \emph{top2} faster position on \emph{14} datasets out of \emph{25}.

\section{Conclusion}
\label{conclusion}
\emph{CIAMS} is a scalable and extensible method for automatic model selection using the clustering indices estimated for a given dataset.  We build an end-to-end pipeline for recommending the best classification model class for a given production dataset based on the dataset's characteristics as represented in the clustering index feature space. Our experimental setup with \emph{60} different binary class datasets confirms the validity of our hypothesis that the classification performance of a dataset is a function of the dataset clustering indices with $R^2>80\% $ score for all the model classes included in the setup. We also observe that our mapper module predicts the expected classification performance within~\emph{10\%} error margin for an average of \emph{two-thirds} of \emph{60} datasets in the subsampling mode. While evaluating the rank-order prediction, we observe that our automatic model selection method scores precise \emph{top3} predictions for \emph{three-fourths} of \emph{60} datasets.  We also develop an end-to-end automatic machine learning system for data classification. A user can send a test dataset and acquire the classification labels without worrying about the classification model selection and building processes. When we compare against popular commercial and open-source automatic machine learning platforms with another set of \emph{25} binary class datasets, we outperform others with an average rank of \emph{1.68} in classification performance, even in the absence of the explicit feature engineering performed by other platforms. Regarding running time, we show that \emph{CIAMS} is significantly faster than the other methods. The next step for \emph{CIAMS} is to extend the platform for multi-class classification and regression tasks to make it a complete Automated ML suite. The codebase for this work is available in Github\footnote{https://github.com/BUDDI-AI/CIAMS}.

\section {Data Availability}
The datasets generated during and/or analyzed during the current study are available from the corresponding author upon reasonable request.

\section{Acknowledgment}
This work was supported by Claritrics Inc. d.b.a BUDDI AI under Grant number RB1920CS200BUDD008156.  On behalf of all authors, the corresponding author states that there is no conflict of interest.

\bibliography{automl}

\end{document}